\documentclass[sigconf,acmthm=false]{acmart}
\setcopyright{none}
\acmDOI{}
\acmISBN{}
\acmConference[HSCC+ICCPS 2027]{HSCC+ICCPS 2027}{May 17--20, 2027}{Boulder, CO, USA}
\acmYear{2027}
\usepackage{graphicx} 
\usepackage{algorithm}

\usepackage[T1]{fontenc}
\usepackage{multirow}
\usepackage{bigdelim}
\usepackage{makecell}
\usepackage{amsmath}
\usepackage{amsthm}
\usepackage{placeins}
\usepackage{algpseudocode}
\usepackage{float}
\usepackage{graphicx}
\usepackage{bm}
\usepackage{mathtools}
\usepackage{physics}
\usepackage{xcolor}
\newif\ifrevisionhighlight
\revisionhighlighttrue
\definecolor{revisionblue}{RGB}{0,70,180}

\usepackage{enumitem}
\usepackage{amsthm}

\usepackage{dsfont}
\usepackage{booktabs}
\newtheorem{definition}{Definition}

\usepackage{newfloat}
\usepackage{listings}
\DeclareCaptionStyle{ruled}{labelfont=normalfont,labelsep=colon,strut=off} 
\floatstyle{ruled}
\newfloat{listing}{tb}{lst}{}
\floatname{listing}{Listing}

\usepackage{booktabs}

\title{Action-Guided Reachability for Deterministic World Models \\of Vision-Based Systems}

\author{Yuang Geng}
\affiliation{
  \institution{University of Florida}
  \city{Gainesville}
  \state{Florida}
  \country{USA}
}

\author{Zhongzheng Zhang}
\affiliation{
  \institution{University of Florida}
  \city{Gainesville}
  \state{Florida}
  \country{USA}
}

\author{Chengzhen Jiang}
\affiliation{
  \institution{University of Florida}
  \city{Gainesville}
  \state{Florida}
  \country{USA}
}

\author{Yanru Li}
\affiliation{
  \institution{University of Florida}
  \city{Gainesville}
  \state{Florida}
  \country{USA}
}

\author{Xinyang Wang}
\affiliation{
  \institution{University of Florida}
  \city{Gainesville}
  \state{Florida}
  \country{USA}
}

\author{Zhuoyang Zhou}
\affiliation{
  \institution{University of Florida}
  \city{Gainesville}
  \state{Florida}
  \country{USA}
}

\author{Ivan Ruchkin}
\affiliation{
  \institution{University of Florida}
  \city{Gainesville}
  \state{Florida}
  \country{USA}
}

\begin{document}

\begin{abstract}
End-to-end vision-based controllers map raw camera images directly to control actions, yet verifying their closed-loop behavior remains challenging: no analytical model maps a physical state to a camera image, and propagating a set of states through high-dimensional images yields loose reachable sets.
We address both challenges with two techniques.
First, we propose a \textit{Deterministic World Model (DWM)}, a latent-free decoder that generates camera images only from the physical state, trained with a saliency-weighted reconstruction loss and a control-consistency loss.
The DWM replaces the camera in the closed loop, so a sound and tight reachable tube of the surrogate closed loop can be computed without bounding any latent variables.
Second, we propose an \textit{action-guided partition-and-verify} scheme that certifies as many initial sets as possible within a fixed computation budget.
A Gaussian process learned from cheap simulations orders the cells and splits those that lie across the property boundary, while every other unknown cell is split along the state dimension that most changes the controller's actions along the simulated closed loop, i.e., the one that most reduces the pixel uncertainty reaching the controller.
On a CARLA braking system and three Gym benchmarks, the DWM is more accurate to the real camera than a cGAN surrogate and yields $1.9\times$ to $69\times$ narrower robustness intervals. 
The vision part causes most of the remaining overapproximation, and our splitting reduces the unknown fraction of the four-dimensional CartPole initial set from $21.88\%$ to $8.18\%$ within the budget.
\end{abstract}

\maketitle


\section{Introduction}
\label{sec:intro}


Modern autonomous systems increasingly make decisions from high-dimensional observations (e.g., camera images and LiDAR) rather than from a state vector~\cite{chen2024end,kroemer2021review}.
These observations carry far richer information about the environment (e.g., obstacle appearance) for control~\cite{tampuu2020survey}.
Among these modalities, vision is the most widely adopted: end-to-end vision-based controllers take raw images as input and generate control actions directly~\cite{tampuu2020survey,levine2016end}.
The camera renders an image of the current state, the controller maps that image to a control action, and the plant dynamics advance the system to the next state~\cite{geng2024bridging}.
These systems are now deployed in safety-critical applications such as autonomous vehicles~\cite{2021selfdriving,2021realcontrolcar}, robotic manipulation~\cite{2023robotCNN}, and drone navigation~\cite{2024IROSdrone}, where a single unsafe action can lead to a collision or a crash.

The safety of these vision-based systems demands rigorous formal guarantees~\cite{2024Vision_Verify_Survey}.
Formal verification of neural network controllers has made substantial progress when the controller input is a low-dimensional state vector (e.g., position and velocity): 
tools such as Verisig~\cite{ivanov2019verisig}, POLAR~\cite{wang2023polar}, and ReachNN~\cite{huang2019reachnn} propagate sets of states through the controller and plant dynamics, step by step, to certify safety properties.
However, none of these tools can be directly applied when the controller input is a \textit{high-dimensional camera image}, due to two fundamental challenges.

\noindent\textbf{Challenge 1: No analytical state-to-image model for complex visual environments.}
Closed-loop verification requires every component in the loop to be analyzable.
For the camera, however, no interpretable equation maps a physical state to a pixel image.
A learned surrogate can replace the camera and close the loop, but it is a black box without physical meaning, and whether it is accurate and verifiable depends on how it is designed.

\noindent\textbf{Challenge 2: High-dimensional images make reachable sets loose and expensive to tighten.}
Even when a state-to-image mapping exists, propagating a set of states through a high-dimensional image layer introduces an overapproximation error at every layer. 
As a result, verification faces a trade-off between precision and cost: analyzing a large initial set yields reachable sets too loose, while analyzing it at a small cell multiplies the computation, especially for higher-dimensional spaces.

We mitigate these two challenges through two complementary techniques: a \textbf{Deterministic World Model (DWM)} that closes the verification loop (Challenge~1), and an \textbf{action-guided partition-and-verify} scheme that keeps the reachable sets tight within a fixed computation budget (Challenge~2).

\looseness=-1
Previous work did replace the camera with a \emph{generative surrogate} inside the verification loop.
In particular, conditional GANs (cGANs) synthesize images conditioned on the system state~\cite{katz2022verification,2025scalableCNN}.
These cGANs also take a stochastic \emph{latent vector}, a random noise sampled from a fixed prior (e.g., Gaussian or uniform), which is introduced to improve image diversity.
However, we posit that the latent vector is harmful for verification: it carries no physical meaning, so its bounds must be chosen wide, which produces an image set far larger than the real camera ever generates and directly exacerbates Challenge~2~\cite{2025scalableCNN}.
Therefore, our DWM removes the latent vector and generates camera images \textit{only} from the physical state.
This deterministic state-to-image mapping closes the verification loop and avoids the spurious image variability introduced by the latent vector.

Splitting the initial set into smaller cells is a standard way to tighten reachable sets, e.g., branch-and-bound~\cite{bunel2020branch,wang2018reluval} or simulation-guided partitioning~\cite{xiang2021reachable,everett2021reachability}.
A cell may need splitting for two reasons.
First, a \emph{mixed cell} contains both satisfying and violating initial states, so no reachable tube can certify property satisfaction. This cell must be split along the target property boundary, which we locate with a Gaussian process fitted to the robustness.
Second, a \emph{loose cell} may satisfy the property, but its tube is too wide to be certified.
For a loose cell in a classical state-based system, the controller takes the state directly, so splitting the widest dimension directly reduces the uncertainty.
However, in a vision-based closed loop, this choice is no longer straightforward: the controller sees pixels rather than states.
We found that many cells are loose because of the camera: a wide cell produces a wide range of images, i.e., a large pixel uncertainty, which the controller turns into loose action bounds.
Meanwhile, each state dimension affects the range of images differently.
Therefore, to tighten a loose cell, we search DWM simulation for the state dimension whose split most reduces the pixel uncertainty (and thus the action discrepancy). 


We evaluate both techniques on four vision-based control benchmarks: a braking system in CARLA and three OpenAI Gym environments (CartPole, MountainCar, and Pendulum).
The DWM yields tighter reachable sets than cGAN surrogates, and the action-guided split is the only rule that certifies $90\%$ of the four-dimensional CartPole initial set within the budget, reducing the unknown fraction from $21.88\%$ to $8.18\%$ compared with bisecting every dimension.
The contributions of this paper are:
\begin{enumerate}
    \item \textbf{Verification-oriented deterministic world model.} A latent-free state-to-image model trained with saliency-weighted reconstruction and control-consistency objectives, which replaces the camera in the verification loop.
    \item \textbf{Action-guided splitting for vision-based Reachability.} A scheme that splits mixed cells along the property boundary located by a Gaussian process, and loose cells along the state dimension that most changes the controller's actions.
    \item \textbf{Systematic experimental evaluation.} On four vision-based control benchmarks, we evaluate world-model fidelity, tube tightness, and certified coverage under a fixed budget, with ablations on each component.
\end{enumerate}


\section{Related Work}
\label{sec:related}

\noindent
\paragraph{Open- and Closed-loop  Verification}

\textit{Open-loop verification} establishes the static input-output relationship of a network without any dynamic feedback~\cite{tran2023verification}.
\textit{Closed-loop verification} treats the network as a controller inside a dynamical system and asks whether the closed-loop trajectory satisfies a safety property~\cite{rossi2024closed_loop,ivanov2019verisig,2021verif_control}.
Tools such as NNV~\cite{lopez2019verification}, POLAR~\cite{wang2023polar}, ReachNN~\cite{huang2019reachnn}, and Verisig~\cite{ivanov2019verisig} have made substantial progress.
However, all of these tools assume the controller's input is a low-dimensional state vector.
When the input is a high-dimensional image, the two challenges from the introduction arise. 

\noindent
\looseness=-1
\paragraph{Closed-loop Verification with Surrogate Models}
Since cameras lack analytical models, several works replace the camera with a \emph{surrogate model} inside the verification loop~\cite{2024Vision_Verify_Survey}. Symbolic abstractions~\cite{santa2022nnlander,habeeb2023cameramodel_math} derive geometry-based bounds on the image set; they work well in structured settings (e.g., runway landing or lane keeping) but struggle in unstructured environments. Generative surrogates instead learn the camera~\cite{katz2022verification,2025scalableCNN}: Katz et al.~\shortcite{katz2022verification} use a cGAN conditioned on the state and a Gaussian latent vector, and Cai et al.~\shortcite{2025scalableCNN} compose the dynamics, cGAN, and controller to reduce one-step error. Both retain stochastic latent variables, which must be bounded during verification; because these variables carry no physical meaning, the bounds are set conservatively wide, forcing the verifier to consider a far larger image set than the system ever produces and causing reachable sets to explode. The root cause is a goal mismatch: latent variables are designed to encourage generative diversity~\cite{isola2017image,zhu2017multimodel_generation}, which directly conflicts with the verification need for tight, physically grounded image sets.

\noindent
\paragraph{Input Splitting for Verification}
Splitting the input set is a standard way to tighten overapproximations.
In open-loop verification, branch-and-bound methods split the input domain or the ReLU activations~\cite{bunel2020branch}; ReluVal bisects the input dimension with the largest smear, i.e., the input-output gradient times the input width~\cite{wang2018reluval}.
In closed-loop verification, the initial state set is partitioned to tighten the reachable sets~\cite{everett2021efficient}, and simulations can guide this partitioning~\cite{xiang2021reachable}.
Sample-based methods with Gaussian processes and Bayesian optimization search for property violations~\cite{ghosh2018verifying,deshmukh2017testing}; they model where the property changes, but not the overapproximation error of a verifier.
None of these rules accounts for a camera surrogate inside the closed loop, where the error comes from the image layer at every step, and a single bound computation is a full multi-step tube.
In contrast, our split rule selects the dimension whose split most separates the actions of the controller, i.e., the action discrepancy between the two halves, which we estimate from a few closed-loop simulations.
Prior works also label cells as unknown, e.g., indeterminate boxes in interval analysis~\cite{jaulin1993set} and uncertain subsets in parameter synthesis~\cite{donze2009parameter}, but do not distinguish mixed cells from loose ones.

\section{Problem Formulation}
\label{background}

\subsection{World Modeling}

Consider a physical plant with state $s_t\in S\subseteq\mathbb R^n$. A camera observes the plant and produces an image $I_t\in\mathcal I\subseteq\mathbb R^{h\times w}$. A neural controller produces an action $u_t\in U\subseteq\mathbb R^m$ that advances the plant.

\begin{definition}[Vision-based Closed-loop Neural System]
\label{def:system}
A vision-based closed-loop neural system is the tuple $(S,\mathcal I,U,P,C,f)$, where the camera $P:S\to\mathcal I$ maps states to images, the neural controller $C:\mathcal I\to U$ maps images to actions, and the dynamics $f:S\times U\to S$ advances the plant. Its evolution is
\begin{equation}
\label{eq:sys-evol}
s_{t+1}=f\bigl(s_t,C(P(s_t))\bigr).
\end{equation}
\end{definition}
Given $s_0$, the trajectory over $T$ steps is $\tau(s_0,T)=[s_0,\ldots,s_T]$, obtained by iterating Eq.~\eqref{eq:sys-evol}.

Verification requires a state-to-image mapping through which sets of states can be propagated, which the real camera does not provide. Thus, \textbf{our first problem} is to learn a camera surrogate $\hat P:S\to\mathcal I$ that replaces $P$ in the loop.
Since the closed loop observes the camera only through the controller, $\hat P$ needs to match $P$ only in what the controller responds to, i.e., $C\circ\hat P\approx C\circ P$ rather than $\hat P\approx P$ pixel by pixel. Formally, we seek
\begin{equation}
    \arg\min_{\hat P}\ \mathbb E_{s\sim\mathcal U(S)}\bigl\|C(\hat P(s))-C(P(s))\bigr\|,
    \label{eq:problem1}
\end{equation}
where $\mathcal U(S)$ is the uniform distribution over $S$.

Replacing $P$ with $\hat P$ while keeping $C$ and $f$ fixed defines the \textit{surrogate system}:
\begin{equation}
\hat s_{t+1}=f\bigl(\hat s_t,C(\hat P(\hat s_t))\bigr),
\qquad \hat s_0=s_0.
\end{equation}
We denote its trajectory by $\tau_{\rm wm}(s_0,T)$. Reachability analysis will be performed on this surrogate closed loop. 

\subsection{Property Verification on Surrogate System}
\label{subsec:problem2}
We verify the surrogate system against a given property over a bounded initial set $S_0\subseteq S$ and a horizon $T$.
The property is a Signal Temporal Logic (STL) formula $\varphi$ over the surrogate trajectory $\tau_{\rm wm}(s_0,T)=[\hat s_0,\ldots,\hat s_T]$, where the hat marks the state of the surrogate system rather than of the real one.
Its robustness $\rho_\varphi(\tau_{\rm wm}(s_0,T))$~\cite{donze2010robust}, written $\rho_\varphi(s_0)$ for short, is nonnegative if and only if the trajectory satisfies $\varphi$, and its magnitude measures the margin of satisfaction or violation.

\noindent
\textbf{Verifier.}
Since $S_0$ contains infinitely many initial states, we verify it subset by subset.
For a subset $\mathcal X\subseteq S_0$, a verifier first computes a sequence of sound reachable sets $\{\mathcal R_t^{\rm wm}(\mathcal X)\}_{t=0}^{T}$ satisfying
\begin{equation}
\label{eq:subpro2}
\hat s_t\in\mathcal R_t^{\rm wm}(\mathcal X)
\quad\text{for all }s_0\in\mathcal X,\ t\in\{0,\ldots,T\},
\end{equation}

Next, the verifier bounds the robustness of every trajectory in this tube.
The robustness bounds of $\mathcal X$ are the extreme robustness over all state trajectories $\tau_{\rm wm}$ that stay in the tube $\mathcal T(\mathcal X)=\mathcal R_0^{\rm wm}(\mathcal X)\times\cdots\times\mathcal R_T^{\rm wm}(\mathcal X)$,
\begin{equation}
\underline\rho(\mathcal X)=\min_{\tau_{\rm wm}\in\mathcal T(\mathcal X)}\rho_\varphi(\tau_{\rm wm}),
\qquad
\overline\rho(\mathcal X)=\max_{\tau_{\rm wm}\in\mathcal T(\mathcal X)}\rho_\varphi(\tau_{\rm wm}).
\label{eq:rho_tube}
\end{equation}
Every surrogate trajectory from $\mathcal X$ stays in the tube by Eq.~\eqref{eq:subpro2}, so $\underline\rho(\mathcal X)\le\rho_\varphi(s_0)\le\overline\rho(\mathcal X)$ for all $s_0\in\mathcal X$.
The \textit{verdict} of the cell is
\begin{equation}
V(\mathcal X)=\left\{
\begin{array}{@{}l@{\quad}l@{}l}
\text{satisfying}, & \underline\rho(\mathcal X)\ge0, & \rdelim\}{2}{3mm}[\ \text{certified}]\\
\text{violating}, & \overline\rho(\mathcal X)<0, & \\
\text{unknown}, & \underline\rho(\mathcal X)<0\le\overline\rho(\mathcal X), &
\end{array}\right.
\label{eq:verdict}
\end{equation}
i.e., $\mathcal X$ is \emph{certified} if $V(\mathcal X)\neq$ unknown.
A cell that is not verified within the budget $B$ also counts as unknown.
The budget $B$ is a wall-clock time limit on a CPU that covers all computation, i.e., the simulations, the model fitting, and the verifier calls.

\noindent
\textbf{Certifying the initial set.}
The high-dimensional intermediate images loosen the tube over multiple steps, so a large cell often remains unknown, whereas many small cells amplify the verification cost.
Thus, \textbf{our second problem} is: given $S_0$, $\varphi$, $T$, and a computation budget $B$, find a partition $\Pi$ of $S_0$ and certify the maximal relative volume of the cells:
\begin{equation}
    \begin{aligned}
    &\max_\Pi \ \operatorname{vol}\bigl(\bigcup\{\mathcal X\in\Pi:V(\mathcal X)\neq\text{unknown}\}\bigr)\\
    &\ \text{s.t. the total verification time}\le B.
    \end{aligned}
    \label{eq:problem2}
\end{equation}

After verification, a cell may remain unknown for two reasons.
It may be \emph{mixed}, i.e., it contains both satisfying and violating initial states, so no tube can certify it as a whole and hence it must be split until each part lies on one side of $\rho_\varphi=0$.
Alternatively, it may be \emph{loose}, i.e., all its states satisfy (or violate) $\varphi$, but the tube is overly inflated so that the robustness bounds still contain zero.
Mixed cells occur in any system, but vision-based systems are particularly affected by loose cells: a wide cell yields a wide set of images, which loosens the actions and thus the tube.
Our intuition is that the two kinds of cells call for different splits: a mixed cell should be split across the property boundary, whereas a loose cell should be split along the state dimension that most changes the actions of the controller.
We address the first problem, i.e., training a faithful surrogate, in Section~\ref{sec:approach}, and the second, i.e., partitioning and certifying $S_0$, in Section~\ref{sec:partition}.

\begin{figure*}[t]
 \begin{center}
\includegraphics[width=\linewidth, trim={0 118 0 5}, clip]{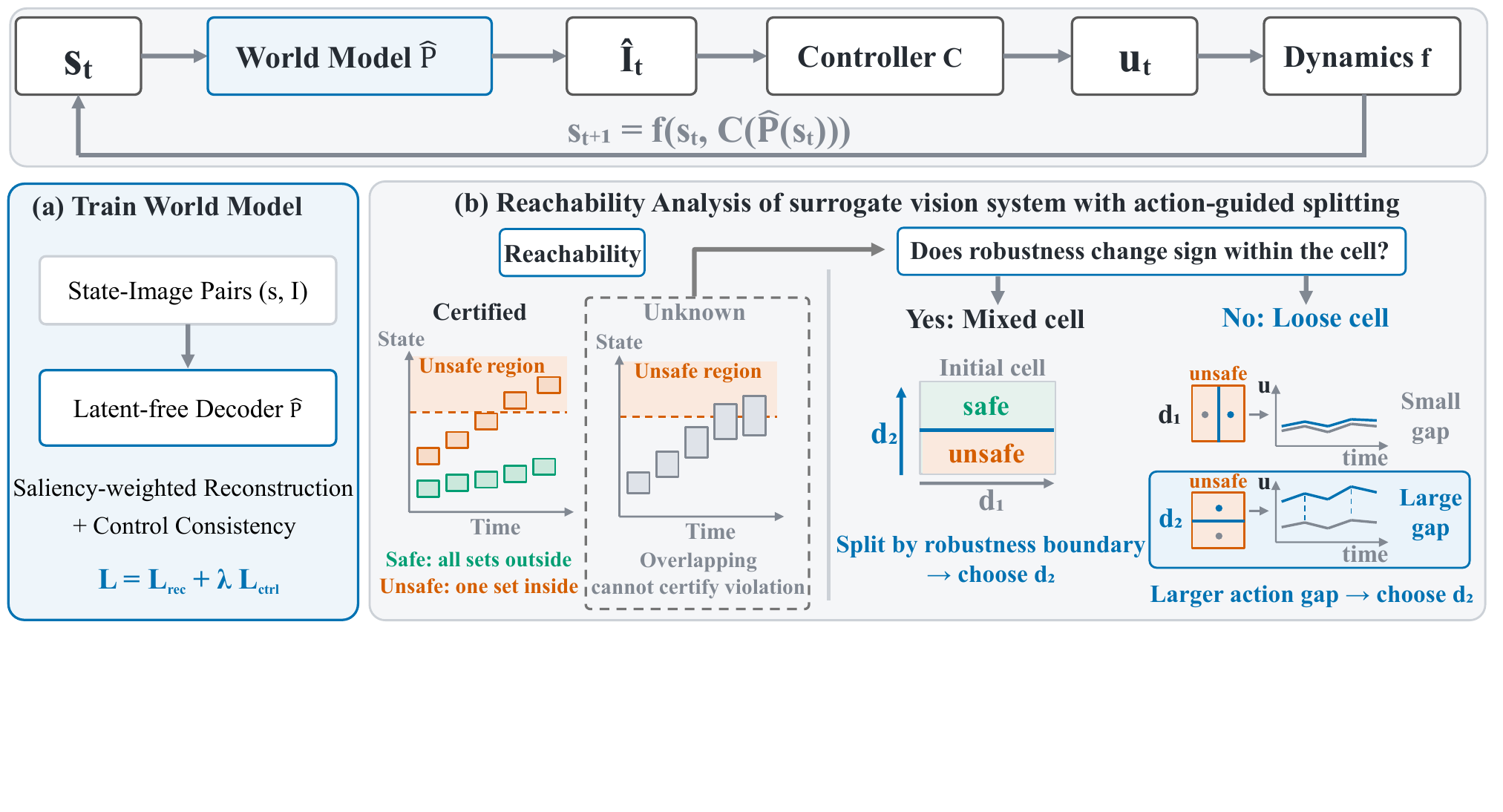}
 
 \caption{Overview of our framework. Top: the DWM $\hat P$ replaces the camera in the closed loop. (a) Training the DWM. (b) Certifying a cell with a sound reachable tube, and splitting an unknown cell: across the predicted robustness boundary if it is mixed, and along the dimension with the larger action gap if it is loose.}
 \label{fig: verification_pipline}
 \end{center}
 \vspace{-4mm}
\end{figure*} 
\section{Training Deterministic World Model}

\label{sec:approach}

This section addresses the first problem by training a deterministic world model, as shown in Figure~\ref{fig: verification_pipline}. 

\paragraph{Dataset}
Our world model learns to approximate the camera module $P(s)$ using the dataset $D_{\text{train}} = \{(s_i, I_i)\}_{i=1}^N$. The dataset does not depend on the initial set $S_0$ or the property $\varphi$, so one world model can be used for any verification query. In practice, we sample states uniformly from $S$, matching the distribution $\mathcal U(S)$ in Eq.~\eqref{eq:problem1}, and record the corresponding camera images.

\paragraph{Model Architecture}
Our world model is implemented with a state-to-image decoder $\hat P: S \to \mathcal{I}$ with parameters $\theta$, which takes a physical state $s$ and outputs a reconstructed image $\hat{I} = \hat P(s)$. 
The visual output has high dimensions $H \times W$ (e.g., $96 \times 96$ grayscale). The architecture follows a typical design, with fully connected layers followed by transposed convolution (a.k.a. deconvolution) layers.

\subsection{Loss Function} 
Our world model is trained with a weighted sum of two complementary losses: a saliency-weighted reconstruction loss and a control consistency loss. The latter loss minimizes the action discrepancy in Eq.~\eqref{eq:problem1} directly. Since many images yield the same action, the reconstruction loss serves as its dense proxy, anchoring the generated image to the real one on the pixels that affect the controller.

\paragraph{Saliency-Weighted Reconstruction Loss:}
A state has only a few degrees of freedom, whereas the image has tens of thousands of pixels. Asking the model to reproduce every pixel accurately from such a low-dimensional input is neither achievable nor necessary: most of the image is background that the controller ignores, and spending capacity on it degrades the reconstruction of the few regions that actually drive the control action. Hence, we intentionally sacrifice global fidelity to improve fidelity in control-critical regions by assigning greater weights to controller-relevant pixels.

Concretely, we assign each pixel $(i,j)$ a saliency score
$\psi_{ij}\in[0,1]$ based on how much occluding its neighborhood changes the
controller's action. We slide a $k\times k$ patch over the image and index the
patch positions by $q=1,\ldots,Q$. For each patch $q$, we replace its pixels
with zero (white under the image convention used here),
leaving the rest of the image unchanged:
\begin{equation}
    I^{(q)}_{ij}=\begin{cases}
        0, & (i,j)\text{ lies in patch }q,\\
        I_{ij}, & \text{otherwise.}
    \end{cases}
    \label{eq:occluded}
\end{equation}
Since neighboring patches overlap, each pixel may be covered by multiple
patches. Its saliency score is the average action change caused by those patches:
\begin{equation}
    H_{ij}=\frac{1}{N_{ij}}
    \sum_{\substack{q:\,\text{patch }q\\\text{covers pixel }(i,j)}}
    \bigl|C(I^{(q)})-C(I)\bigr|.
    \label{eq:occlusion_heat}
\end{equation}
Here, $N_{ij}$ is the number of patches covering pixel $(i,j)$.
A larger $H_{ij}$ means that occluding the pixel's neighborhood has a larger
influence on the action. We normalize the scores per image:
\begin{equation}
    \psi_{ij}=\frac{H_{ij}-\min H}{\max H-\min H},
    \label{eq:saliency}
\end{equation}
where $\min H$ and $\max H$ are taken over all pixels in that image. If all
scores are equal, we set $\psi_{ij}=0$.
The weighted mean squared error over the $H\times W$ image is

\begin{equation}
    L_{\mathrm{rec}}(I,\hat{I})
    =
    \frac{1}{HW}
    \sum_{i=1}^{H}\sum_{j=1}^{W}
    w_{ij}\bigl(\hat{I}_{ij}-I_{ij}\bigr)^2,
    \label{eq:weighted_mse}
\end{equation}
\begin{equation}
    w_{ij}
    =
    w_l + (w_h-w_l)\psi_{ij}.
    \label{eq:pixel_weight}
\end{equation}
so that control-relevant pixels ($\psi_{ij}\!\to\!1$) receive high weight ($w_h$) and control-irrelevant background ($\psi_{ij}\!\to\!0$) receives low weight ($w_l$). Tying the weights to the controller's occlusion responses makes our training independent of any assumption about how the objects of interest appear.

\paragraph{Control Consistency Loss:}
The control consistency loss ensures that the reconstructed image produces a similar control behavior to the original image. 

Recall that per Definition~\ref{def:system}, $C(\cdot)$ is the image-based controller that maps an image to a control action $u = C(I)$. Inspired by a similar loss for image-based control repair~\cite{sobolewski2025generalizable}, we penalize the discrepancy between the control actions derived from the reconstructed image and the original image:
\begin{equation}
    L_{\mathrm{ctrl}}(I, \hat{I})
    = \bigl\|\, C(\hat{I}) - C(I) \,\bigr\|_2^2.
    \label{eq:controller_loss}
\end{equation}

\medskip
The final world model training loss $L(\theta)$ combines the two above into a weighted sum:  
\begin{equation}
    L(\theta)
    = L_{\mathrm{rec}}(I, \hat{I})
    + \lambda L_{\mathrm{ctrl}}(I, \hat{I}),
    \label{eq:total_loss}
\end{equation}
The hyperparameter $\lambda$ balances image and action fidelity; we choose $\lambda\in[0.1,1]$, since a larger $\lambda$ degrades the image reconstruction while the action error has already converged (see Section~\ref{subsec:ablation}). 

\section{Action-Guided Partition-and-Verify}
\label{sec:partition}

This section addresses the second problem: soundly certifying the property as satisfied or violated over as much of the initial-state space $S_0$ as possible within a computation budget $B$ for the surrogate vision-based closed-loop system.
The size of the initial cells creates a trade-off: cells that are too large yield loose reachable tubes and often remain unknown, whereas cells that are too small require many verifier calls, including in regions that a larger cell could already certify.

Therefore, we will refine unknown cells adaptively, according to why they remain unknown.
After constructing sound tubes (Section~\ref{subsec:tubes}), we use cheap simulations to separate mixed cells from loose ones and to prioritize verifier calls (Section~\ref{subsec:gp}), and we split loose cells along the dimension through which the camera loosens the tube most (Section~\ref{subsec:split}).

\subsection{Sound Reachable Tubes}
\label{subsec:tubes}
For each initial cell, we construct a reachable tube that contains all the trajectories of the surrogate system starting from the cell, using linear relaxation~\cite{zhang2018efficient,singh2019abstract,xu2020automatic}.
Given a cell $\mathcal X=[\underline s,\overline s]\subseteq S_0$, we start from $\mathcal R_0=\mathcal X$ and propagate a box $\mathcal R_t\supseteq\{\hat s_t : \hat s_0\in\mathcal X\}$ step by step.
At step $t$, we bound the action $C(\hat P(\hat s_t))$ over $\mathcal R_t$ by linear relaxation, and propagate $\mathcal R_t$ and the action bound through the interval extension of the dynamics $f$ to obtain $\mathcal R_{t+1}$.
We use the forward+backward mode of auto\_LiRPA~\cite{xu2020automatic}, which avoids a backward pass for every neuron of the large image layer of $\hat P$; other verification backends such as IBP, backward CROWN, and StarV are compared in Section~\ref{subsec:backends}.

\textbf{Reachable tube robustness.}
We compute the bounds of Eq.~\eqref{eq:rho_tube} for any STL formula with its interval semantics.
STL robustness combines the values of atomic predicates by $\min$ and $\max$~\cite{donze2010robust}, so bounding each predicate over the box $\mathcal R_t$ and applying the formula to the lower and upper ends soundly gives $\underline\rho(\mathcal X)$ and $\overline\rho(\mathcal X)$.
For example, for the property $\varphi$ of Section~\ref{subsec:problem2}, the Braking System requires the vehicle never to collide and to stop no farther than $d_{\max}$ from the leading vehicle,
\begin{equation}
\begin{aligned}
&\varphi=\square_{[0,T]}(d_t\geq0)\wedge(d_T\leq d_{\max}),\\
&\rho_\varphi(s_0)=\min\bigl(\min\nolimits_{t}d_t,\ d_{\max}-d_T\bigr).
\end{aligned}
\label{eq:stl_robustness}
\end{equation}
which gives
\begin{equation}
\begin{aligned}
\underline\rho(\mathcal X)&=\min\bigl(\min\nolimits_{t}\underline d_t,\ d_{\max}-\overline d_T\bigr),\\
\overline\rho(\mathcal X)&=\min\bigl(\min\nolimits_{t}\overline d_t,\ d_{\max}-\underline d_T\bigr),
\end{aligned}
\label{eq:tube_robustness}
\end{equation}
where $[\underline d_t,\overline d_t]$ is the range of $d$ in $\mathcal R_t$.
The verdict of the cell then follows from these bounds as in Section~\ref{subsec:problem2}.


\subsection{Sample-guided Splitting and Ordering}
\label{subsec:gp}
Every verifier call computes a full multi-step tube, so verifier calls dominate the cost of partition-and-verify.
However, the tube of an unknown cell does not tell whether the cell is mixed of safe and unsafe states or the tube is too loose (Section~\ref{subsec:problem2}).
Therefore, we learn an approximation of the robustness $\rho_\varphi(s_0)$ over $S_0$ from simulations of the surrogate closed loop, and its zero level set predicts the property boundary.
We use this model to split mixed cells across the predicted boundary and to decide which cells to verify first.

\paragraph{Robustness model.}
Before verification, we simulate $1{,}000$ surrogate trajectories from initial states sampled uniformly in $S_0$ and fit a Gaussian process (GP)~\cite{rasmussen2006gaussian} to their robustness $\rho_\varphi(s_0)$.
GP models of the robustness are widely used to search for property violations of cyber-physical systems~\cite{deshmukh2017testing,ghosh2018verifying}, and GP-based level set estimation locates where an unknown function crosses a threshold from few samples~\cite{bryan2005active,gotovos2013active}; we use the similar model to locate the boundary $\rho_\varphi=0$, but only to guide a sound verifier. The GP provides mean $\mu(s)$ and standard deviation $\sigma(s)$ and stays fixed afterwards. 

\paragraph{Splitting mixed cells.}
We regard an unknown cell as mixed if the GP mean $\mu$ changes sign over this cell.
No tube can certify a mixed cell as a whole, so splitting it only pays off if it separates the satisfying part from the violating part.
We therefore cut it along the dimension in which $\mu$ varies most over this cell, i.e., across the predicted boundary.
Every other unknown cell is regarded as loose and is split by the action-guided rule of Section~\ref{subsec:split}.

\paragraph{Queue order.}
Certified volume should accumulate early within the budget, so each cell is ranked by the volume it is expected to certify, i.e., its volume weighted by how likely it is to be certified:
\begin{equation}
    p(\mathcal X)=\underbrace{|\mathcal X|}_{\text{volume}}\cdot\underbrace{q(\mathcal X)}_{\text{certifiability}},
    \qquad
    q(\mathcal X)=\min\left\{\frac{\min_{s\in \mathcal X}|\mu(s)|}{W_{\rm p}},1\right\},
    \label{eq:priority}
\end{equation}
where $|\mathcal X|$ is the volume of the cell and $W_{\rm p}=\overline\rho(\mathcal X^{\rm p})-\underline\rho(\mathcal X^{\rm p})$ is the width of the robustness bounds of its parent cell $\mathcal X^{\rm p}$, and cells are processed in decreasing order of $p$.

The factor $q$ compares the predicted robustness margin with how loose the tube is; since the tube of a new cell is not yet computed, the parent width $W_{\rm p}$ estimates its looseness.
We set $q=1$ for $S_0$ itself, and a cell whose parent was skipped uses the width of its nearest verified ancestor.

\paragraph{Skipping.}
The GP also skips verifier calls that would be wasted: a cell in which the GP is uncertain about the sign, i.e., $|\mu(s)|\leq\sigma(s)$, is split without a call. A mixed cell at the minimum width is marked unknown without a call.
Every verdict still comes from the sound tube of Section~\ref{subsec:tubes}, and the GP only decides how cells are split, ordered, and skipped. Therefore, the certified cells remain sound regardless of the accuracy of the GP.

\subsection{Action-Guided Splitting}
\label{subsec:split}
A loose cell needs a different splitting strategy than a tight cell. All its states satisfy (or violate) $\varphi$, so its tube is unknown only because of the overapproximation error from reachability.
The GP models the true robustness and provides no information about this error: cutting a loose cell along the dimension in which $\mu$ varies most does not reduce the overapproximation error.

In a vision-based closed loop, this error comes mainly from the camera surrogate. 
A wide cell yields a wide set of images, i.e., a large pixel uncertainty, which the linear relaxation of $C\circ\hat P$ turns into loose action bounds, and the dynamics accumulates these bounds over the horizon.
Thus, a good split is the one that most reduces the pixel uncertainty reaching the controller.
For example, the safety of CartPole hardly depends on the cart velocity, yet the velocity moves the cart in the next frames and thereby changes the images and actions along the whole horizon.

Unlike a state-feedback controller, the camera observes only part of the state, so a one-step sensitivity of the action, e.g., the input gradient used for state-feedback networks~\cite{wang2018reluval}, vanishes on such hidden dimensions.
Therefore, our split is \emph{action-guided}, and we call it action-guided splitting (AGS): we implicitly manage the pixel uncertainty of each state dimension through the actions that the controller computes from the surrogate images in the closed loop.

\paragraph{Action-divergence score.}
For each dimension $d$ of a loose cell, we cut the cell in half along dimension $d$ and simulate the surrogate closed-loop trajectory from the center of each half, giving the trajectories $\{\hat s_t^{(1)}\}$ and $\{\hat s_t^{(2)}\}$. The action-divergence score of $d$ is the discrepancy between the actions along the two trajectories,
\begin{equation}
    \mathcal S_d(\mathcal X)=\sum_{t=0}^{T-1}\bigl\|C(\hat P(\hat s_t^{(1)}))-C(\hat P(\hat s_t^{(2)}))\bigr\|_1 ,
    \label{eq:action_score}
\end{equation}
and we cut along the dimension $d^*$ with the largest score $S_d$.
A large score means the two separate cells drive the controller differently, so the cut removes much of the action range, and thus of the relaxation error.
Since the actions are collected along the whole trajectory, the score also captures state variables that the camera does not see directly but that reach the image through the dynamics.

\section{Experimental Setup}
\label{sec:experiment}

This section presents the systems, neural controllers, reachability setup, and baselines for comparison.

\noindent
\paragraph{Systems}
We evaluate our world model using four dynamics commonly used in vision-based verification: \textit{CartPole}, \textit{MountainCar},  \textit{Pendulum}, and \textit{Braking System}.
CartPole, MountainCar, and Pendulum follow the dynamics of OpenAI Gym~\cite{brockman2016openai}, with $\Delta t=0.02$\,s for CartPole and $0.05$\,s for Pendulum.
The Braking System is a vehicle approaching a stopped one, with $d_{t+1}=d_t-\Delta t\,v_t$ and $v_{t+1}=\max(0,\,v_t-\Delta t\,(k_{\rm brake}u_t+a_0))$, where $u_t\in[0,1]$ is the brake command and $\Delta t=0.1$\,s.

\noindent
\paragraph{Neural Controllers}
The controller has two convolutional layers and two fully connected layers.
Given an input image $I_t$, either a real camera frame or a synthesized world-model image, the controller outputs the control input $u_t = C(I_t)$.
We train four controllers using reinforcement learning, one for each dynamic. Each controller is trained to optimize a task reward (e.g., keeping the pole upright, reaching the hilltop, or avoiding collision); the goal sets above are independently defined verification specifications.
It has two convolutional layers ($1\to4\to1$ channels, kernel $4$, stride $2$) and two dense layers ($576\to64\to1$) with ReLU, and a sigmoid (CartPole, Braking System) or tanh (MountainCar, Pendulum) output.
The DWM decoder $\hat P$ maps the two camera-relevant state variables through three dense layers ($32$, $64$, $432$ units) to a $3\times12\times12$ tensor and through three transposed convolutions ($3\to4\to8\to1$ channels, kernel $4$, stride $2$) to a $96\times96$ image, with a sigmoid output (a clamp to $[0,1]$ for the Braking System).

\noindent
\paragraph{Initial Sets and Properties}
Table~\ref{tab:initial_sets} lists the initial set $S_0$ and the STL property $\varphi$ of each benchmark, with horizon $T=20$.
The camera sees only $(x,\theta)$ of CartPole and $(d,v)$ of the Braking System; the experiments on fixed boxes (Sections~\ref{subsec:dwm_cgan}--\ref{subsec:backends}) vary only these two variables, and partition-and-verify certifies the whole 4-D $S_0$.

\begin{table}[t]
\centering
\caption{Initial sets $S_0$ and properties $\varphi$ ($T=20$).}
\label{tab:initial_sets}
\small
\setlength{\tabcolsep}{2pt}
\begin{tabular}{llll}
\toprule
Benchmark & State & $S_0$ & Property $\varphi$ \\
\midrule
CartPole & $(x,\dot x,\theta,\dot\theta)$ & \makecell[l]{$[0,0.6]\times[-0.2,0.2]$\\$\times[0.06,0.12]\times[-0.2,0.2]$} & $\square_{[0,T]}|\theta_t|\le12^\circ$ \\
\addlinespace
MountainCar & $(x,\dot x)$ & $[-0.6,0.45]\times[0,0.07]$ & $\lozenge_{[0,T]}x_t\ge0.45$ \\
\addlinespace
Pendulum & $(\theta,\dot\theta)$ & $[1,2]\times[4.5,5]$ & $\lozenge_{[0,T]}|\theta_t|\le0.15$ \\
\addlinespace
Braking & \makecell[l]{$(d,v,$\\$\ k_{\rm brake},a_0)$} & \makecell[l]{$[6,8]\times[6,6.4]$\\$\times[8,10]\times[0.3,0.55]$} & \makecell[l]{$\square_{[0,T]}(d_t\ge0)$\\$\wedge(d_T\le3.5)$} \\
\bottomrule
\end{tabular}
\end{table}

\noindent
\paragraph{Baselines}
We compare our DWM against a generative surrogate camera with a latent variable.
Following prior work~\cite{2025scalableCNN}, we replace our DWM decoder with a conditional GAN whose generator takes the concatenation of the state $s_t$ and a two-dimensional latent vector $z \sim \mathcal{U}([-r,r]^2)$ and outputs an image $\tilde I_t = \mathrm{Gan}(s_t,z)$. During reachability analysis, both $s$ and $z$ must be bounded; we bound $z$ by the range used in training, i.e., $r=0.8$ for CartPole, MountainCar, and Pendulum, and $r=0.05$ for the Braking System.

\noindent
\paragraph{Ablation and Sensitivity}
To isolate the two design choices of the DWM (removing the latent and adding $L_{\mathrm{ctrl}}$), we compare three surrogates: the cGAN baseline, the DWM trained with $L_{\mathrm{rec}}$ only ($\lambda=0$ in Eq.~\eqref{eq:total_loss}), and the full DWM.
We measure each by the metrics of Table~\ref{tab:dwm_cgan} (Section~\ref{subsec:ablation}).
To analyze the sensitivity to the loss weight, we further vary $\lambda\in\{0,0.01,0.1,0.5,1,10,100\}$ with all other settings fixed, and report the image MSE, the action error, and the trajectory discrepancy.

\noindent
\paragraph{Verification Tools}
We bound $C\circ\hat P$ with four tools: IBP~\cite{gowal2018effectiveness}, LiRPA (forward+backward)~\cite{xu2020automatic}, backward CROWN~\cite{zhang2018efficient}, and approximate star sets in StarV~\cite{tran2025starv}.
All other parts of the reachability analysis are shared, and LiRPA is the default tool in all other experiments.

\noindent
\paragraph{Metrics}
We measure fidelity to the real camera by the image MSE and the action error $\lvert C(\hat I)-C(I)\rvert$ on the held-out test split, and by the trajectory discrepancy, i.e., the mean absolute state error between the real and the surrogate closed loop from $20$ initial states, normalized by the range of each state variable.
For the Braking System, the real frame at each step is the dataset frame with the nearest distance.
cGAN errors are averaged over eight latent draws.
We measure verifiability by verifying both surrogates with LiRPA (forward+backward) on the same ten initial boxes: three bands of $\rho_\varphi$ (satisfying, near the boundary, violating) with boxes of $1\%$, $3\%$, and $10\%$ of the width of $S_0$, plus one $3\%$ box at a random center.
For the cGAN, the latent is an additional input interval over its training range.
We report the width of the robustness interval, the tube area projected onto two state variables and averaged over the steps, and the time per box.

\noindent
\paragraph{Partition-and-Verify Setup}
We certify the whole $S_0$ of each benchmark (Table~\ref{tab:initial_sets}) with the partition-and-verify loop of Section~\ref{sec:partition}, down to a minimum width of $2^{-8}$ of $S_0$ per dimension ($2^{-6}$ for 4-D).
In Braking System~(4-D), the brake gain $k_{\rm brake}$ and the base deceleration $a_0$ are constants.

\noindent
\paragraph{Partition Baselines and Metrics}
\textit{Uniform} verifies cells from the largest to the smallest and bisects every free dimension, with IBP or LiRPA.
The other methods use LiRPA with the GP ordering and skipping of Section~\ref{subsec:gp} and differ only in the split:
\textit{GP + all dims} bisects every free dimension, \textit{GP + widest dim} the widest one relative to $S_0$, \textit{GP + GP-variance dim} the one in which $\mu$ varies most (the mixed-cell rule applied to every cell), and \textit{AGS (ours)} uses the action-guided split of Section~\ref{subsec:split}.
We also adapt two splitting rules from the literature. \textit{GP + smear dim} keeps the GP ordering and splits the dimension with the largest smear $|\partial C(\hat P(s))/\partial s_d|$ times the cell width $w_d$ at the cell center, as ReluVal does for open-loop networks~\cite{wang2018reluval}.
\textit{GSG} is the greedy simulation-guided partitioning of~\cite{everett2021reachability}: it uses no GP, verifies first the cell whose parent tube exceeds the hull of the trajectories sampled from the parent the most, and bisects the widest dimension, since~\cite{everett2021reachability} does not fix the split dimension.
Every method runs with the same budget $B=400$\,s on a CPU ($800$\,s for 4-D), which covers the simulations, the GP fitting, and the verifier calls, and every run stops at $B$.
We report $\mathrm{AUC}=\frac{1}{B}\int_0^B C(t)\,dt$, where $C(t)$ is the certified fraction of $S_0$ at time $t$, the time $T_{90\%}$ to certify $90\%$ of $S_0$, the number of verifier calls, and the unknown fraction, all measured at $B$.

\section{Results Analysis}
\label{sec:results}

We first show how the latent of the cGAN loosens verification and compare the DWM with the cGAN (Section~\ref{subsec:dwm_cgan}), ablate the two design choices of the DWM, and analyze its sensitivity to $\lambda$ (Section~\ref{subsec:ablation}).
We then compare verification tools and locate the source of the overapproximation (Section~\ref{subsec:backends}), and evaluate splitting techniques (Section~\ref{subsec:partition}).

\subsection{Comparison with Baseline Surrogate}
\label{subsec:dwm_cgan}

\begin{figure}[h]
 \begin{center}
 \includegraphics[width=\linewidth, trim={100 80 40 0}, clip]{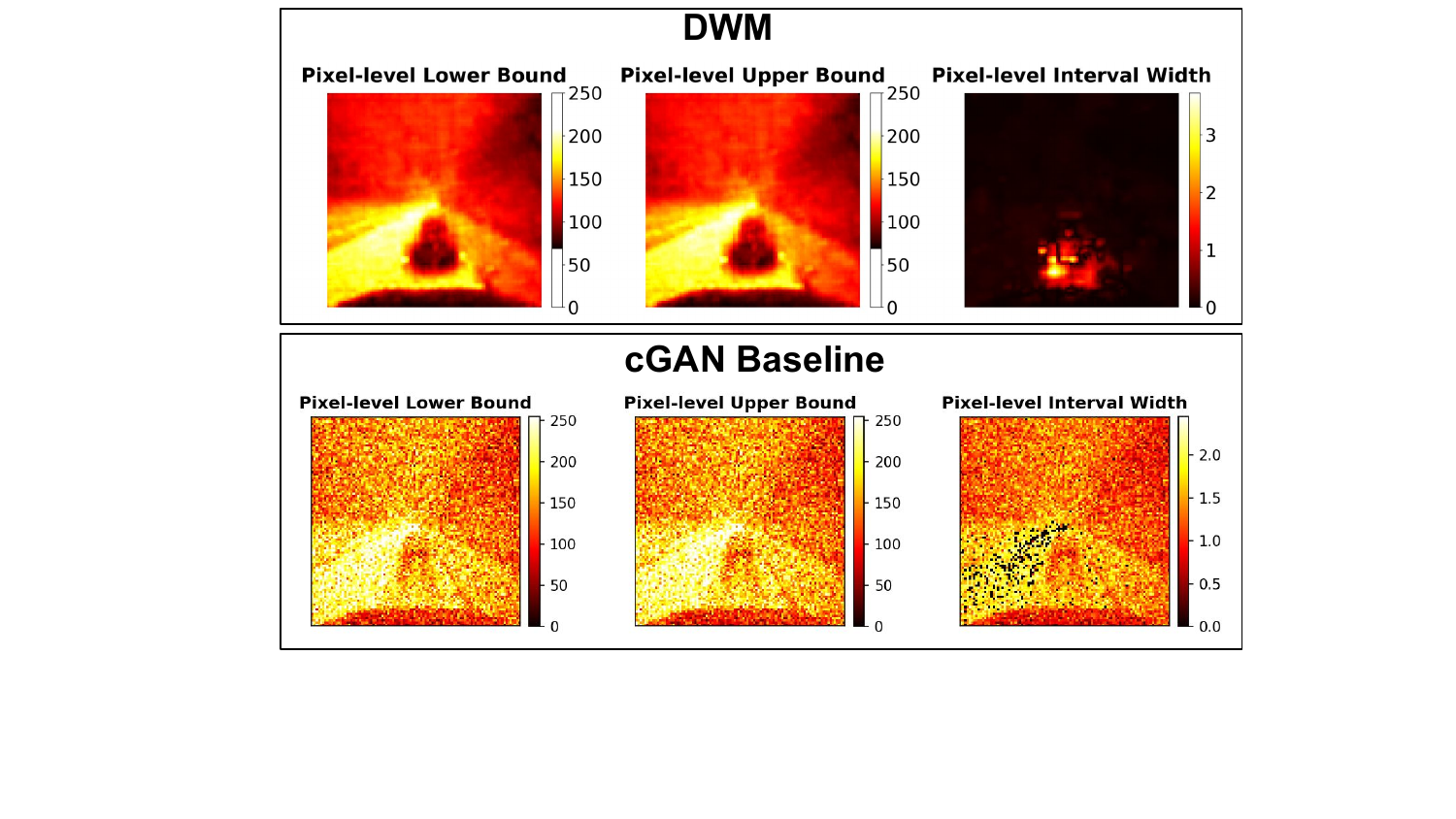}
\caption{Pixel-level bounds of the DWM (top) and the cGAN (bottom) on the Braking System for the same state interval $[6.20,6.21]\times[5.20,5.21]$; the cGAN latent $z\in[-0.08,0.08]^2$.}
\label{fig: carla_compare}
\end{center}
\end{figure}

\textbf{Generated Images Suffer from Uninterpretable Latents.}
The latent of the cGAN widens the image set far beyond what the camera produces, which loosens the reachable tube.
Visual comparisons in Figure~\ref{fig: carla_compare} show that, for the same state interval, the cGAN produces significantly larger pixel-level intervals than our DWM.
Uncontrolled latent variables $z$ inject noise into the background regions that are irrelevant to control, further inflating the pixel interval.
When these widened pixel intervals propagate through the controller and plant dynamics over multiple steps, overapproximation errors accumulate rapidly.

\begin{table*}[t]
\centering
\caption{DWM vs.\ cGAN, with the DWM trained without $\mathcal L_{\rm ctrl}$ as ablation (Section~\ref{subsec:ablation}). Image and action errors on the test split; trajectory discrepancy to the real closed loop in \% of the state range. Reachability: mean $\pm$ std over ten initial boxes, with the tube-area scale under each benchmark. Lower is better.}
\label{tab:dwm_cgan}
\small
\begin{tabular}{ll ccc ccc}
\toprule
\multirow{2}{*}{Benchmark} & \multirow{2}{*}{Model}
& \multicolumn{3}{c}{Fidelity to the real camera}
& \multicolumn{3}{c}{Reachability (10 boxes)} \\
\cmidrule(lr){3-5}\cmidrule(lr){6-8}
& & Image MSE & Action MAE & Traj.\ disc.\ (\%) & $\overline\rho-\underline\rho$ & Tube area & Time / box (s) \\
\midrule
\multirow{3}{*}{\makecell[l]{CartPole\\(area $\times10^{-3}$)}}
& cGAN & 0.0273 & 0.0641 & 5.49 $\pm$ 0.98 & 0.963 $\pm$ 0.038 & 220.5 $\pm$ 10.9 & 1.18 $\pm$ 0.03 \\
& DWM w/o $\mathcal L_{\rm ctrl}$ & \textbf{0.0019} & 0.0382 & \textbf{0.46} $\pm$ 0.52 & 0.015 $\pm$ 0.018 & \textbf{0.35} $\pm$ 0.52 & \textbf{0.67} $\pm$ 0.05 \\
& DWM  & \textbf{0.0019} & \textbf{0.0136} & 0.80 $\pm$ 0.80 & \textbf{0.014} $\pm$ 0.015 & \textbf{0.35} $\pm$ 0.50 & 0.68 $\pm$ 0.04 \\
\midrule
\multirow{3}{*}{\makecell[l]{MountainCar\\(area $\times10^{-3}$)}}
& cGAN & 0.0480 & 0.3109 & 3.52 $\pm$ 2.35 & 0.322 $\pm$ 0.101 & 16.09 $\pm$ 6.73 & 1.21 $\pm$ 0.03 \\
& DWM w/o $\mathcal L_{\rm ctrl}$ & \textbf{0.0023} & 0.2483 & 3.05 $\pm$ 0.51 & 0.181 $\pm$ 0.176 & 8.44 $\pm$ 9.97 & 0.66 $\pm$ 0.04 \\
& DWM  & 0.0029 & \textbf{0.0184} & \textbf{0.07} $\pm$ 0.04 & \textbf{0.173} $\pm$ 0.181 & \textbf{8.10} $\pm$ 10.03 & \textbf{0.64} $\pm$ 0.05 \\
\midrule
\multirow{3}{*}{\makecell[l]{Pendulum\\(area $\times1$)}}
& cGAN & 0.0076 & 0.0201 & 0.13 $\pm$ 0.08 & 1.533 $\pm$ 0.230 & 30.20 $\pm$ 5.38 & 1.22 $\pm$ 0.02 \\
& DWM w/o $\mathcal L_{\rm ctrl}$ & \textbf{0.0008} & 0.0176 & 0.12 $\pm$ 0.07 & 0.812 $\pm$ 0.611 & \textbf{5.94} $\pm$ 7.68 & 0.65 $\pm$ 0.06 \\
& DWM  & \textbf{0.0008} & \textbf{0.0125} & \textbf{0.08} $\pm$ 0.04 & \textbf{0.793} $\pm$ 0.603 & 6.13 $\pm$ 8.52 & \textbf{0.64} $\pm$ 0.09 \\
\midrule
\multirow{3}{*}{\makecell[l]{Braking System\\(area $\times10^{-2}$)}}
& cGAN & 0.0254 & 0.1226 & 5.26 $\pm$ 1.14 & 1.832 $\pm$ 0.137 & 75.79 $\pm$ 11.97 & 1.16 $\pm$ 0.03 \\
& DWM w/o $\mathcal L_{\rm ctrl}$ & \textbf{0.0077} & 0.0259 & 5.28 $\pm$ 1.35 & 0.568 $\pm$ 0.592 & 14.83 $\pm$ 23.91 & 0.69 $\pm$ 0.06 \\
& DWM  & 0.0080 & \textbf{0.0254} & \textbf{4.45} $\pm$ 1.27 & \textbf{0.477} $\pm$ 0.622 & \textbf{13.64} $\pm$ 24.11 & \textbf{0.65} $\pm$ 0.06 \\
\bottomrule
\end{tabular}
\end{table*}

\noindent
\textbf{Fidelity and verifiability.}
On every benchmark, the DWM is both more faithful to the real camera and much easier to verify than the cGAN (Table~\ref{tab:dwm_cgan}).
Its image MSE is $3.2\times$ to $16.6\times$ lower and its action error $1.6\times$ to $16.9\times$ lower.
These errors do not accumulate over the horizon: the DWM trajectories deviate from the real ones by $0.07\%$ to $0.8\%$ of the state range on CartPole, MountainCar, and Pendulum, $1.7\times$ to $50\times$ less than the cGAN, and by $4.45\%$ vs.\ $5.26\%$ on the Braking System.
The gap widens in reachability: under the same reachability tool, the cGAN yields $1.9\times$ to $69\times$ wider robustness intervals and $2.0\times$ to $637\times$ larger tubes, and it is about $1.8\times$ slower to verify because of its uninterpretable latent.
The large standard deviations of the DWM come from the nested box widths, whereas the cGAN tube stays wide even for the $1\%$ boxes, as Figure~\ref{fig: carla_compare} suggests.

\subsection{Ablation and Sensitivity of the DWM}
\label{subsec:ablation}
\textbf{Ablation of the latent and $\mathcal L_{\rm ctrl}$.}
Removing the latent makes the surrogate tight, and adding $\mathcal L_{\rm ctrl}$ makes it faithful in the closed loop (Table~\ref{tab:dwm_cgan}: cGAN vs.\ DWM w/o $\mathcal L_{\rm ctrl}$, and DWM w/o $\mathcal L_{\rm ctrl}$ vs.\ DWM).
Removing the latent lowers the image MSE by $3.3\times$ to $20.9\times$ and narrows the robustness interval on all four benchmarks, e.g., from $0.963$ to $0.015$ on CartPole.
A low image error alone, however, does not make the controller behave as on real images: without $\mathcal L_{\rm ctrl}$, the DWM trajectories deviate from the real ones by $3.05\%$ on MountainCar and $5.28\%$ on the Braking System, close to the cGAN ($3.52\%$ and $5.26\%$).
Adding $\mathcal L_{\rm ctrl}$ reduces the action error on all four benchmarks, most on MountainCar ($0.248\to0.018$), and the trajectory discrepancy on MountainCar ($3.05\%\to0.07\%$), Pendulum ($0.12\%\to0.08\%$), and the Braking System ($5.28\%\to4.45\%$); on CartPole, both variants stay below $1\%$ of the state range.
Therefore, the control consistency loss is needed for the closed-loop fidelity of the DWM.

\noindent
\textbf{Sensitivity to $\lambda$.}
We found that a moderate $\lambda\in[0.1,1]$ keeps both the action and the image accurate, whereas a larger $\lambda$ barely improves the action but makes the image highly inaccurate (Table~\ref{tab:lambda}, all other settings fixed).
Adding $L_{\mathrm{ctrl}}$ reduces the action error sharply, e.g., from $0.249$ to $0.020$ on MountainCar and from $0.040$ to $0.013$ on CartPole at $\lambda=0.1$, while the image MSE for $\lambda\in[0.1,1]$ stays within $1.8\times$ that of $\lambda=0$.
Beyond $\lambda=1$, the action error barely changes (e.g., $0.016\to0.013$ on MountainCar from $\lambda=1$ to $100$, and $0.010\to0.014$ on Pendulum), whereas the image MSE grows monotonically, to $2.8\times$--$10.8\times$ that of $\lambda=0$ at $\lambda=100$, since the control term dominates the reconstruction.
The trajectory discrepancy stays small even at $\lambda=100$, since the controller only reads the pixels that $L_{\mathrm{ctrl}}$ fits; the DWM then fits the given controller rather than the camera.
Therefore, we use $\lambda = 0.1$ in all experiments.

\begin{table}[H]
\centering
\caption{Sensitivity to $\lambda$ (Eq.~\ref{eq:total_loss}): image MSE ($\times10^{-3}$) and action MAE ($\times10^{-2}$) on the test split, and trajectory discrepancy to the real closed loop.}
\label{tab:lambda}
\normalsize
\setlength{\tabcolsep}{3pt}
\begin{tabular}{ll ccccccc}
\toprule
& $\lambda$ & $0$ & $0.01$ & $0.1$ & $0.5$ & $1$ & $10$ & $100$ \\
\midrule
\multirow{3}{*}{CartPole} & Image & \textbf{1.9} & 2.9 & \textbf{1.9} & 2.8 & 3.5 & 7.0 & 16.3 \\
 & Action & 3.97 & 1.28 & 1.30 & \textbf{0.95} & 1.04 & 1.05 & 1.03 \\
 & Traj.\ (\%) & 1.45 & 1.11 & 1.19 & \textbf{1.02} & 1.04 & 0.51 & 1.04 \\
\midrule
\multirow{3}{*}{\makecell[l]{Mountain\\Car}} & Image & \textbf{2.3} & 2.5 & 2.7 & 2.9 & 3.3 & 5.0 & 14.3 \\
 & Action & 24.89 & 6.21 & 2.03 & 1.82 & 1.60 & 1.36 & \textbf{1.32} \\
 & Traj.\ (\%) & 3.07 & 0.62 & 0.11 & 0.07 & 0.07 & \textbf{0.06} & \textbf{0.06} \\
\midrule
\multirow{3}{*}{Pendulum} & Image & 0.8 & 0.7 & \textbf{0.6} & 0.8 & 0.8 & 3.0 & 8.1 \\
 & Action & 1.81 & 1.73 & 1.29 & 1.23 & \textbf{0.99} & 1.23 & 1.35 \\
 & Traj.\ (\%) & 0.10 & 0.10 & 0.11 & 0.08 & \textbf{0.07} & 0.09 & 0.12 \\
\midrule
\multirow{3}{*}{Braking} & Image & \textbf{7.6} & 7.7 & 7.8 & 8.2 & 8.2 & 10.7 & 21.1 \\
 & Action & 2.65 & 2.59 & 2.54 & 2.49 & \textbf{2.45} & 2.52 & 2.46 \\
 & Traj.\ (\%) & 3.87 & 3.90 & 3.65 & 2.58 & \textbf{1.92} & 2.54 & \textbf{1.92} \\
\bottomrule
\end{tabular}
\end{table}

\subsection{Tightness of Reachable Tubes}
\label{subsec:backends}
\textbf{Verification backends.}
The DWM does not depend on the verifier, so we compare four backends on the DWM closed loop: IBP, LiRPA (forward+backward)~\cite{xu2020automatic}, backward CROWN~\cite{zhang2018efficient}, and approximate Star-set reachability (StarV).
Only the bound on $C\circ\hat P$ differs, and each of the ten boxes of Section~\ref{subsec:dwm_cgan} is verified by a single call.

\begin{figure*}[h]
\centering
\includegraphics[width=0.9\textwidth]{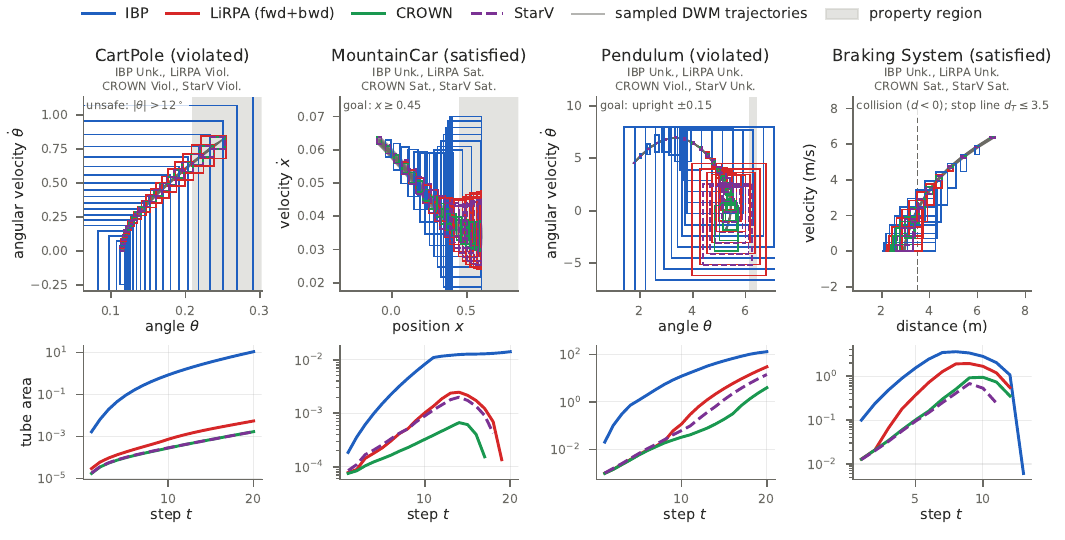}
\caption{\looseness=-1
Top: tubes of the four backends from one initial box per benchmark (CartPole and Pendulum: violating; others: satisfying), with $100$ sampled DWM trajectories. Bottom: tube area per step (log scale).}
\label{fig:backend_tubes}
\end{figure*}

\begin{table}[t]
\centering
\caption{Verification techniques on the DWM closed loop, mean $\pm$ std over the ten boxes of Table~\ref{tab:dwm_cgan}, without any splitting. Lower is better.}
\label{tab:backends}
\footnotesize
\setlength{\tabcolsep}{2pt}
\begin{tabular}{ll ccc}
\toprule
Benchmark & Backend & $\overline\rho-\underline\rho$ & Tube area & Time / box (s) \\
\midrule
\multirow{4}{*}{\makecell[l]{CartPole\\(area $\times10^{-3}$)}}
& IBP & 0.695 $\pm$ 0.252 & 144.52 $\pm$ 70.52 & \textbf{0.07} $\pm$ 0.00 \\
& LiRPA (fwd+bwd) & 0.013 $\pm$ 0.013 & 0.32 $\pm$ 0.42 & 0.81 $\pm$ 0.02 \\
& CROWN & \textbf{0.010} $\pm$ 0.009 & 0.24 $\pm$ 0.31 & 52.87 $\pm$ 3.00 \\
& StarV & \textbf{0.010} $\pm$ 0.008 & \textbf{0.23} $\pm$ 0.30 & 893.65 $\pm$ 242.61 \\
\midrule
\multirow{4}{*}{\makecell[l]{MountainCar\\(area $\times10^{-3}$)}}
& IBP & 0.315 $\pm$ 0.108 & 15.52 $\pm$ 6.92 & \textbf{0.08} $\pm$ 0.00 \\
& LiRPA (fwd+bwd) & 0.171 $\pm$ 0.181 & 7.90 $\pm$ 10.13 & 1.11 $\pm$ 0.08 \\
& CROWN & \textbf{0.143} $\pm$ 0.158 & \textbf{5.69} $\pm$ 8.67 & 70.88 $\pm$ 4.69 \\
& StarV & 0.161 $\pm$ 0.166 & 6.84 $\pm$ 8.82 & 767.23 $\pm$ 77.56 \\
\midrule
\multirow{4}{*}{\makecell[l]{Pendulum\\(area $\times1$)}}
& IBP & 1.533 $\pm$ 0.230 & 28.72 $\pm$ 5.67 & \textbf{0.10} $\pm$ 0.00 \\
& LiRPA (fwd+bwd) & 0.847 $\pm$ 0.610 & 6.92 $\pm$ 8.62 & 1.07 $\pm$ 0.02 \\
& CROWN & \textbf{0.545} $\pm$ 0.471 & \textbf{2.48} $\pm$ 3.67 & 74.49 $\pm$ 9.78 \\
& StarV & 0.571 $\pm$ 0.549 & 2.90 $\pm$ 5.24 & 372.57 $\pm$ 89.80 \\
\midrule
\multirow{4}{*}{\makecell[l]{Braking System\\(area $\times10^{-2}$)}}
& IBP & 1.960 $\pm$ 0.119 & 86.77 $\pm$ 11.03 & \textbf{0.13} $\pm$ 0.04 \\
& LiRPA (fwd+bwd) & 0.624 $\pm$ 0.523 & 14.50 $\pm$ 17.40 & 1.85 $\pm$ 0.29 \\
& CROWN & 0.231 $\pm$ 0.270 & 2.58 $\pm$ 6.10 & 97.69 $\pm$ 49.46 \\
& StarV & \textbf{0.205} $\pm$ 0.212 & \textbf{1.69} $\pm$ 3.62 & 149.29 $\pm$ 39.32 \\
\bottomrule
\end{tabular}
\end{table}

\begin{table*}[!t]
\centering
\caption{Partition-and-verify over $S_0$ with the same budget $B=400$\,s ($800$\,s for 4-D) for all methods, mean $\pm$ std over three seeds. All metrics are measured at $B$. $T_{90\%}$: time to certify $90\%$ of $S_0$; $>B$: not reached within $B$; --: never reached. Sim.: simulation time within $B$.}
\label{tab:partition}
\footnotesize
\setlength{\tabcolsep}{4pt}
\begin{tabular}{ll ccccc}
\toprule
Benchmark & Method & AUC $\uparrow$ & $T_{90\%}$ (s) $\downarrow$ & Verifier calls $\downarrow$ & Unknown (\%) $\downarrow$ & Sim. (s) \\
\midrule
\multirow{8}{*}{MountainCar}
& Uniform + IBP        & 0.711 & -- & 29{,}125 & 27.49 & -- \\
& Uniform + LiRPA      & 0.926 $\pm$ 0.000 & 76 $\pm$ 0 & 3{,}837 $\pm$ 0 & \textbf{0.94} $\pm$ 0.00 & -- \\
& GP + all dims        & \textbf{0.928} $\pm$ 0.001 & \textbf{63} $\pm$ 0 & 2{,}998 $\pm$ 26 & 1.01 $\pm$ 0.01 & 1.4 $\pm$ 0.0 \\
& GP + widest dim      & 0.916 $\pm$ 0.001 & 75 $\pm$ 1 & 3{,}028 $\pm$ 29 & 1.01 $\pm$ 0.01 & 1.5 $\pm$ 0.0 \\
& GP + GP-variance dim & 0.898 $\pm$ 0.005 & 101 $\pm$ 5 & 3{,}746 $\pm$ 54 & 1.01 $\pm$ 0.01 & 1.5 $\pm$ 0.0 \\
& GP + smear dim~\cite{wang2018reluval} & 0.768 $\pm$ 0.001 & 211 $\pm$ 2 & 5{,}575 $\pm$ 38 & 1.01 $\pm$ 0.01 & 1.5 $\pm$ 0.0 \\
& GSG~\cite{everett2021reachability} & 0.898 $\pm$ 0.001 & 100 $\pm$ 1 & 4{,}567 $\pm$ 0 & \textbf{0.94} $\pm$ 0.00 & 24.5 $\pm$ 0.3 \\
& AGS (ours)           & 0.919 $\pm$ 0.001 & 68 $\pm$ 1 & \textbf{2{,}780} $\pm$ 29 & 1.01 $\pm$ 0.01 & 8.3 $\pm$ 0.3 \\
\midrule
\multirow{8}{*}{Pendulum}
& Uniform + IBP        & 0.000 & -- & 68{,}437 & 100.00 & -- \\
& Uniform + LiRPA      & 0.395 $\pm$ 0.000 & $>$400 & 7{,}456 $\pm$ 30 & 22.68 $\pm$ 0.05 & -- \\
& GP + all dims        & 0.474 $\pm$ 0.002 & $>$400 & 6{,}992 $\pm$ 40 & 20.44 $\pm$ 0.24 & 1.5 $\pm$ 0.0 \\
& GP + widest dim      & 0.490 $\pm$ 0.003 & $>$400 & 6{,}951 $\pm$ 52 & 12.64 $\pm$ 0.39 & 1.5 $\pm$ 0.0 \\
& GP + GP-variance dim & \textbf{0.662} $\pm$ 0.002 & \textbf{285} $\pm$ 2 & 6{,}830 $\pm$ 84 & \textbf{4.29} $\pm$ 0.06 & 1.5 $\pm$ 0.0 \\
& GP + smear dim~\cite{wang2018reluval} & 0.524 $\pm$ 0.002 & 373 $\pm$ 2 & 6{,}510 $\pm$ 30 & 7.85 $\pm$ 0.14 & 1.5 $\pm$ 0.0 \\
& GSG~\cite{everett2021reachability} & 0.329 $\pm$ 0.002 & $>$400 & 6{,}282 $\pm$ 40 & 37.37 $\pm$ 0.31 & 44.1 $\pm$ 0.5 \\
& AGS (ours)           & \textbf{0.662} $\pm$ 0.002 & 289 $\pm$ 1 & \textbf{6{,}627} $\pm$ 52 & 4.31 $\pm$ 0.09 & 18.2 $\pm$ 0.2 \\
\midrule
\multirow{8}{*}{CartPole (4-D)}
& Uniform + IBP        & 0.000 & -- & 132{,}689 & 100.00 & -- \\
& Uniform + LiRPA      & 0.564 $\pm$ 0.003 & $>$800 & 14{,}481 $\pm$ 283 & 27.46 $\pm$ 0.28 & -- \\
& GP + all dims        & 0.606 $\pm$ 0.001 & $>$800 & 13{,}718 $\pm$ 60 & 21.88 $\pm$ 0.09 & 1.5 $\pm$ 0.0 \\
& GP + widest dim      & 0.662 $\pm$ 0.002 & $>$800 & 13{,}308 $\pm$ 40 & 17.94 $\pm$ 0.10 & 1.5 $\pm$ 0.0 \\
& GP + GP-variance dim & 0.190 $\pm$ 0.003 & $>$800 & 13{,}078 $\pm$ 185 & 63.36 $\pm$ 0.12 & 1.5 $\pm$ 0.0 \\
& GP + smear dim~\cite{wang2018reluval} & 0.033 $\pm$ 0.000 & $>$800 & 12{,}676 $\pm$ 26 & 86.11 $\pm$ 0.11 & 1.4 $\pm$ 0.0 \\
& GSG~\cite{everett2021reachability} & 0.498 $\pm$ 0.000 & $>$800 & 12{,}916 $\pm$ 30 & 33.29 $\pm$ 0.12 & 95.2 $\pm$ 0.1 \\
& AGS (ours)           & \textbf{0.820} $\pm$ 0.002 & \textbf{556} $\pm$ 13 & \textbf{11{,}830} $\pm$ 149 & \textbf{8.18} $\pm$ 0.08 & 70.0 $\pm$ 0.5 \\
\midrule
\multirow{8}{*}{Braking System (4-D)}
& Uniform + IBP        & 0.096 & -- & 135{,}537 & 89.11 & -- \\
& Uniform + LiRPA      & 0.086 $\pm$ 0.001 & $>$800 & 17{,}030 $\pm$ 241 & 87.21 $\pm$ 0.00 & -- \\
& GP + all dims        & 0.093 $\pm$ 0.001 & $>$800 & \textbf{13{,}270} $\pm$ 92 & 80.55 $\pm$ 0.12 & 1.6 $\pm$ 0.0 \\
& GP + widest dim      & 0.110 $\pm$ 0.001 & $>$800 & 15{,}407 $\pm$ 114 & 76.39 $\pm$ 0.19 & 1.6 $\pm$ 0.0 \\
& GP + GP-variance dim & 0.238 $\pm$ 0.001 & $>$800 & 14{,}967 $\pm$ 11 & 60.96 $\pm$ 0.03 & 1.6 $\pm$ 0.0 \\
& GP + smear dim~\cite{wang2018reluval} & 0.081 $\pm$ 0.001 & $>$800 & 14{,}620 $\pm$ 92 & 83.25 $\pm$ 0.00 & 1.6 $\pm$ 0.0 \\
& GSG~\cite{everett2021reachability} & 0.023 $\pm$ 0.004 & $>$800 & 14{,}858 $\pm$ 66 & 96.28 $\pm$ 0.62 & 123.4 $\pm$ 0.1 \\
& AGS (ours)           & \textbf{0.319} $\pm$ 0.002 & $>$800 & 13{,}335 $\pm$ 164 & \textbf{55.55} $\pm$ 0.02 & 112.5 $\pm$ 1.2 \\
\bottomrule
\end{tabular}
\end{table*}

\begin{figure*}[!t]
\centering
\includegraphics[width=\textwidth]{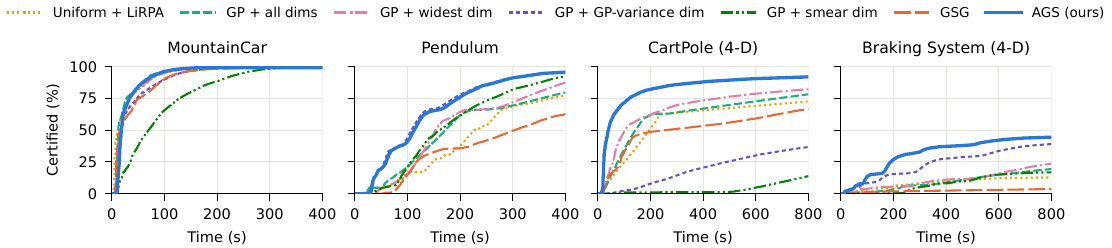}
\caption{Certified fraction $C(t)$ of $S_0$ over the budget for the partition-and-verify methods of Table~\ref{tab:partition}, mean over three seeds.}
\label{fig:certified_curves}
\end{figure*}

\noindent
\textbf{Tightness vs.\ cost.}
LiRPA gives the best trade-off between tightness and cost for partition-and-verify (Table~\ref{tab:backends} and Figure~\ref{fig:backend_tubes}).
IBP is the fastest but the loosest: its robustness interval is $1.8\times$ to $53\times$ wider than that of LiRPA.
LiRPA costs $11\times$ to $14\times$ more than IBP, but still only $1$--$2$\,s per box.
CROWN and StarV are tighter, e.g., $0.231$ and $0.205$ vs.\ $0.624$ on the Braking System, and within $1.3\times$ of LiRPA on CartPole and MountainCar. However, CROWN is $53\times$ to $70\times$ slower than LiRPA. 
Since partition-and-verify needs thousands of calls (Table~\ref{tab:partition}), we use LiRPA, which is $1.8\times$ to $53\times$ tighter than IBP and $53\times$ to $70\times$ faster than CROWN.

\noindent
\textbf{Source of the overapproximation.}
We found that the vision part, not the dynamics, causes most of the overapproximation, and its share grows with the size of the initial box.
Even the tightest reachable tube leaves a gap to the true reachable set, which comes from two places: the linear relaxation of $C\circ\hat P$, i.e., the vision part, and the interval extension of the dynamics $f$.
To understand the source of the overapproximation, we build three tubes from each box over the full horizon.
The \emph{true} tube is the hull of $4{,}096$ DWM trajectories from the box regarded as the true reachable set.
The \emph{dynamics-only} tube replaces the bound on $C\circ\hat P$ by its exact range, i.e., the minimum and maximum action over a dense grid (e.g., $64\times64$) of the decoder-input box, so that only the dynamics overapproximate.
The \emph{LiRPA} tube bounds both the vision part and the dynamics with LiRPA.
With $W$ the tube width averaged over the steps and the state variables (each normalized by its range over the samples), the overapproximation caused by the vision part is
\begin{equation}
    \eta=\frac{W_{\rm LiRPA}-W_{\rm dyn}}{W_{\rm LiRPA}-W_{\rm true}},
    \label{eq:vision_share}
\end{equation}
and the rest, $1-\eta$, is caused by the dynamics.
Since the dynamics-only tube uses a sampled action range, it is not sound and serves only for this analysis.

Table~\ref{tab:error_split} reports $\eta$ for the ten boxes of Table~\ref{tab:dwm_cgan}, grouped by box width.
The vision share is $67\%$ to $78\%$ on average and grows with the box: from $36\%$--$62\%$ on the $1\%$ boxes to $83\%$--$96\%$ on the $10\%$ boxes.
On a small box, the images barely change and the relaxation of $C\circ\hat P$ is almost exact, so the remaining excess comes from the dynamics; on a large box, the pixel uncertainty grows and the relaxation dominates.
The same holds in absolute terms: the dynamics alone widen the sampled tube by a median factor of $1.01$ to $1.47$ on CartPole, MountainCar, and the Braking System, while LiRPA widens it by $1.03\times$ to $7.8\times$.
Pendulum is the exception, where the interval dynamics already widen the tube by $4.6\times$, yet the vision part still adds most of the remaining $18.6\times$.
Hence, the unknown cells that matter, i.e., the large ones, are loose mainly because of the camera, which motivates splitting along the pixel uncertainty (Section~\ref{subsec:split}) rather than tightening the dynamics.

\subsection{Certifying the Whole Initial Set}
\label{subsec:partition}

Our split certifies the largest fraction of $S_0$ on both 4-D benchmarks and stays close to the best rule on the 2-D ones.
We certify the whole $S_0$ of each benchmark with the setup and baselines of Section~\ref{sec:experiment}; Table~\ref{tab:partition} summarizes the runs and Figure~\ref{fig:certified_curves} plots the certified fraction over time.

\paragraph{Backend and ordering.}
As shown in Table~\ref{tab:partition}, a finer partition cannot compensate for a loose backend: IBP certifies no part of $S_0$ on CartPole~(4-D) and Pendulum, and leaves $27.49\%$ of MountainCar unknown, compared with $0.94\%$ for LiRPA.
With the same backend and bisection, the GP ordering certifies more of $S_0$ earlier. On MountainCar, it saves $22\%$ of the calls and certifies $90\%$ of $S_0$ earlier ($63$\,s vs.\ $76$\,s), and on Pendulum, it raises the AUC from $0.395$ to $0.474$.
However, bisecting every dimension still leaves $22\%$ of CartPole~(4-D) unknown at $B$, since each bisection creates $16$ subsets.

\begin{table}[H]
\centering
\caption{Share $\eta$ of the overapproximation error caused by $C\circ\hat P$ (Eq.~\ref{eq:vision_share}), mean $\pm$ std over the boxes of each width; the rest comes from the dynamics. Initial set width: side length of the box as a fraction of the width of $S_0$ in each dimension.}
\label{tab:error_split}
\small
\setlength{\tabcolsep}{4pt}
\begin{tabular}{lcccc}
\toprule
\makecell[l]{Initial set width\\(\% of $S_0$)} & CartPole & MountainCar & Pendulum & Braking \\
\midrule
$1\%$  & $36\%\pm17\%$ & $38\%\pm10\%$ & $46\%\pm27\%$ & $62\%\pm35\%$ \\
$3\%$  & $68\%\pm8\%$ & $92\%\pm6\%$ & $80\%\pm9\%$ & $84\%\pm14\%$ \\
$10\%$ & $96\%\pm1\%$ & $89\%\pm6\%$ & $83\%\pm3\%$ & $84\%\pm15\%$ \\
\midrule
Average    & $67\%\pm26\%$ & $75\%\pm25\%$ & $71\%\pm23\%$ & $78\%\pm25\%$ \\
\bottomrule
\end{tabular}
\end{table}

\paragraph{Action-guided splitting.}
The choice of the split dimension matters most on the 4-D variants.
On CartPole~(4-D), AGS raises the AUC from $0.606$ (GP + all dims) to $0.820$ and lowers the unknown fraction from $21.88\%$ to $8.18\%$. It is the only method that certifies $90\%$ of $S_0$ within $B$ ($556$\,s; all others $>800$\,s).
The property-driven rule collapses on the same benchmark (AUC $0.190$): it splits the cart velocity in only $6\%$ of its splits, since $\rho_\varphi$ hardly depends on it, whereas AGS splits it in $34\%$, since the velocity moves the cart in the next frames.
On Braking System~(4-D), AGS again gives the highest AUC ($0.319$ vs.\ $0.238$) and the lowest unknown fraction ($55.55\%$ vs.\ $60.96\%$), and spends $53\%$ of its splits on the brake gain and base deceleration, which the camera does not see but which change the distance in later frames.
On the two-dimensional benchmarks, AGS matches the best rule on Pendulum (AUC $0.662$ for both, $4.31\%$ vs.\ $4.29\%$ unknown) with $3\%$ fewer calls than GP + GP-variance dim, and certifies $90\%$ of MountainCar almost as early as GP + all dims ($68$\,s vs.\ $63$\,s) with $7\%$ fewer calls.
The two rules from the literature do not transfer to the closed loop.
The smear only measures how the current action depends on each dimension, so GP + smear dim never splits the cart velocity, the brake gain, or the base deceleration, which the current image does not show, and leaves $86.11\%$ of CartPole~(4-D) and $83.25\%$ of Braking System~(4-D) unknown.
GSG comes close to AGS on MountainCar (AUC $0.898$ vs.\ $0.919$), but it still bisects the widest dimension and leaves $33.29\%$ of CartPole~(4-D) unknown, and its simulations take $95$\,s to $123$\,s of the budget on the 4-D benchmarks.
These results indicate that the action-divergence score matters most when the initial set contains dimensions that the property ignores but that affect the camera through the dynamics.

\section{Conclusion}
\label{sec:conclusion}

\looseness=-1

We presented a framework for closed-loop reachability analysis of end-to-end vision-based controllers. A Deterministic World Model (DWM) replaces the camera with a latent-free, state-to-image decoder, trained with a
saliency-weighted reconstruction loss and a control-consistency loss so that it stays faithful where it matters for control. Since every image is governed only by the physical state, a sound reachable tube of the DWM closed loop contains all of its trajectories from an initial set, without bounding any latent variable.
Across four benchmarks, the DWM is more faithful than the cGAN baseline at the image level ($3.2\times$ to $16.6\times$ lower image MSE), the behavior level ($1.6\times$ to $16.9\times$ lower action error), and the trajectory level ($1.2\times$ to $50\times$ lower discrepancy from the real closed loop). From the same initial boxes and under the same verifier, it yields $1.9\times$ to $69\times$ narrower robustness intervals and $2.0\times$ to $637\times$ smaller tubes.

\noindent
\textit{Limitations and future work.}
Our certificates hold only for the surrogate closed loop (not the real system), and the fidelity of the DWM to the real camera is evaluated empirically through the image and action errors.
Transferring the certificates to the real system, e.g., by statistically bounding the trajectory deviation between the two closed loops, is left for future work.
In addition, the DWM may struggle with more complex scenes, such as cluttered backgrounds or large lighting variations.


\bibliographystyle{ACM-Reference-Format}
\bibliography{aaai2027}

@inproceedings{2025scalableCNN,
  title={Scalable surrogate verification of image-based neural network control systems using composition and unrolling},
  author={Cai, Feiyang and Fan, Chuchu and Bak, Stanley},
  booktitle={Proceedings of the AAAI Conference on Artificial Intelligence},
  volume={39},
  number={1},
  pages={21--30},
  year={2025}
}

@INPROCEEDINGS{2024IROSdrone,
  author={Zheng, Haokun and Rajadnya, Sidhant and Zakhor, Avideh},
  booktitle={2024 IEEE/RSJ International Conference on Intelligent Robots and Systems (IROS)}, 
  title={Monocular Depth Estimation for Drone Obstacle Avoidance in Indoor Environments}, 
  year={2024},
  volume={},
  number={},
  pages={10027-10034},
  doi={10.1109/IROS58592.2024.10802577}}

@inproceedings{2021selfdriving,
  title={Maneuver-based trajectory prediction for self-driving cars using spatio-temporal convolutional networks},
  author={Mersch, Benedikt and H{\"o}llen, Thomas and Zhao, Kun and Stachniss, Cyrill and Roscher, Ribana},
  booktitle={2021 IEEE/RSJ International Conference on Intelligent Robots and Systems (IROS)},
  pages={4888--4895},
  year={2021},
  organization={IEEE}
}

@inproceedings{2021realcontrolcar,
  title={Real-time control using convolution neural network for self-driving cars},
  author={Dangskul, Woraphicha and Phattaravatin, Kunanon and Rattanaporn, Kiattisak and Kidjaidure, Yuttana},
  booktitle={2021 7th International Conference on Engineering, Applied Sciences and Technology (ICEAST)},
  pages={125--128},
  year={2021},
  organization={IEEE}
}

@article{2023robotCNN,
  title={Convolutional neural network-based robot control for an eye-in-hand camera},
  author={Guo, Jia and Nguyen, Huu-Thiet and Liu, Chao and Cheah, Chien Chern},
  journal={IEEE Transactions on Systems, Man, and Cybernetics: Systems},
  volume={53},
  number={8},
  pages={4764--4775},
  year={2023},
  publisher={IEEE}
}

@article{huang2019reachnn,
  title={Reachnn: Reachability analysis of neural-network controlled systems},
  author={Huang, Chao and Fan, Jiameng and Li, Wenchao and Chen, Xin and Zhu, Qi},
  journal={ACM Transactions on Embedded Computing Systems (TECS)},
  volume={18},
  number={5s},
  pages={1--22},
  year={2019},
  publisher={ACM New York, NY, USA}
}

@article{wang2023polar,
  title={Polar-express: Efficient and precise formal reachability analysis of neural-network controlled systems},
  author={Wang, Yixuan and Zhou, Weichao and Fan, Jiameng and Wang, Zhilu and Li, Jiajun and Chen, Xin and Huang, Chao and Li, Wenchao and Zhu, Qi},
  journal={IEEE Transactions on Computer-Aided Design of Integrated Circuits and Systems},
  volume={43},
  number={3},
  pages={994--1007},
  year={2023},
  publisher={IEEE}
}

@article{katz2022verification,
  title={Verification of image-based neural network controllers using generative models},
  author={Katz, Sydney M and Corso, Anthony L and Strong, Christopher A and Kochenderfer, Mykel J},
  journal={Journal of Aerospace Information Systems},
  volume={19},
  number={9},
  pages={574--584},
  year={2022},
  publisher={American Institute of Aeronautics and Astronautics}
}

@inproceedings{tran2025starv,
  title={StarV: A Qualitative and Quantitative Verification Tool for Learning-Enabled Systems},
  author={Tran, Hoang-Dung and Choi, Sung Woo and Li, Yuntao and Liu, Qing and Okamoto, Hideki and Hoxha, Bardh and Fainekos, Georgios},
  booktitle={International Conference on Computer Aided Verification},
  pages={376--394},
  year={2025},
  organization={Springer}
}

@inproceedings{geng2024bridging,
  title={Bridging dimensions: Confident reachability for high-dimensional controllers},
  author={Geng, Yuang and Baldauf, Jake Brandon and Dutta, Souradeep and Huang, Chao and Ruchkin, Ivan},
  booktitle={International Symposium on Formal Methods},
  pages={381--402},
  year={2024},
  organization={Springer}
}

@article{singh2019abstract,
  title={An abstract domain for certifying neural networks},
  author={Singh, Gagandeep and Gehr, Timon and P{\"u}schel, Markus and Vechev, Martin},
  journal={Proceedings of the ACM on Programming Languages},
  volume={3},
  number={POPL},
  pages={1--30},
  year={2019},
  publisher={ACM New York, NY, USA}
}

@inproceedings{tran2023verification,
  title={Verification of recurrent neural networks with star reachability},
  author={Tran, Hoang Dung and Choi, Sung Woo and Yang, Xiaodong and Yamaguchi, Tomoya and Hoxha, Bardh and Prokhorov, Danil},
  booktitle={Proceedings of the 26th ACM International Conference on Hybrid Systems: Computation and Control},
  pages={1--13},
  year={2023}
}

@inproceedings{ivanov2019verisig,
  title={Verisig: verifying safety properties of hybrid systems with neural network controllers},
  author={Ivanov, Radoslav and Weimer, James and Alur, Rajeev and Pappas, George J and Lee, Insup},
  booktitle={Proceedings of the 22nd ACM International Conference on Hybrid Systems: Computation and Control},
  pages={169--178},
  year={2019}
}

@inproceedings{lopez2019verification,
  title={Verification of Closed-loop Systems with Neural Network Controllers.},
  author={Lopez, Diego Manzanas and Musau, Patrick and Tran, Hoang-Dung and Johnson, Taylor T},
  booktitle={ARCH@ CPSIoTWeek},
  pages={201--210},
  year={2019}
}

@article{rossi2024closed_loop,
  title={Neural networks in closed-loop systems: Verification using interval arithmetic and formal prover},
  author={Rossi, Federico and Bernardeschi, Cinzia and Cococcioni, Marco},
  journal={Engineering Applications of Artificial Intelligence},
  volume={137},
  pages={109238},
  year={2024},
  publisher={Elsevier}
}

@article{zhu2017multimodel_generation,
  title={Toward multimodal image-to-image translation},
  author={Zhu, Jun-Yan and Zhang, Richard and Pathak, Deepak and Darrell, Trevor and Efros, Alexei A and Wang, Oliver and Shechtman, Eli},
  journal={Advances in neural information processing systems},
  volume={30},
  year={2017}
}

@inproceedings{isola2017image,
  title={Image-to-image translation with conditional adversarial networks},
  author={Isola, Phillip and Zhu, Jun-Yan and Zhou, Tinghui and Efros, Alexei A},
  booktitle={Proceedings of the IEEE conference on computer vision and pattern recognition},
  pages={1125--1134},
  year={2017}
}

@article{habeeb2023cameramodel_math,
  title={Verification of camera-based autonomous systems},
  author={Habeeb, P and Deka, Nabarun and D’Souza, Deepak and Lodaya, Kamal and Prabhakar, Pavithra},
  journal={IEEE Transactions on Computer-Aided Design of Integrated Circuits and Systems},
  volume={42},
  number={10},
  pages={3450--3463},
  year={2023},
  publisher={IEEE}
}

@inproceedings{santa2022nnlander,
  title={Nnlander-verif: A neural network formal verification framework for vision-based autonomous aircraft landing},
  author={Santa Cruz, Ulices and Shoukry, Yasser},
  booktitle={NASA Formal Methods Symposium},
  pages={213--230},
  year={2022},
  organization={Springer}
}

@incollection{2024Vision_Verify_Survey,
  title={Formal Verification Techniques for Vision-Based Autonomous Systems--A Survey},
  author={Mitra, Sayan and P{\u{a}}s{\u{a}}reanu, Corina and Prabhakar, Pavithra and Seshia, Sanjit A and Mangal, Ravi and Li, Yangge and Watson, Christopher and Gopinath, Divya and Yu, Huafeng},
  booktitle={Principles of Verification: Cycling the Probabilistic Landscape: Essays Dedicated to Joost-Pieter Katoen on the Occasion of His 60th Birthday, Part III},
  pages={89--108},
  year={2024},
  publisher={Springer}
}

@INPROCEEDINGS{2021verif_control,
  author={Everett, Michael},
  booktitle={2021 60th IEEE Conference on Decision and Control (CDC)}, 
  title={Neural Network Verification in Control}, 
  year={2021},
  volume={},
  number={},
  pages={6326-6340},
  doi={10.1109/CDC45484.2021.9683154}}

@article{sobolewski2025generalizable,
  title={Generalizable Image Repair for Robust Visual Autonomous Racing},
  author={Sobolewski, Carson and Mao, Zhenjiang and Vejre, Kshitij and Ruchkin, Ivan},
  journal={arXiv preprint arXiv:2503.05911},
  year={2025}
}

@inproceedings{xu2020automatic,
  title={Automatic Perturbation Analysis for Scalable Certified Robustness and Beyond},
  author={Xu, Kaidi and Shi, Zhouxing and Zhang, Huan and Wang, Yihan and Chang, Kai-Wei and Huang, Minlie and Kailkhura, Bhavya and Lin, Xue and Hsieh, Cho-Jui},
  booktitle={Advances in Neural Information Processing Systems},
  volume={33},
  pages={1129--1141},
  year={2020}
}

@inproceedings{donze2010robust,
  title={Robust satisfaction of temporal logic over real-valued signals},
  author={Donz{\'e}, Alexandre and Maler, Oded},
  booktitle={International Conference on Formal Modeling and Analysis of Timed Systems (FORMATS)},
  pages={92--106},
  year={2010},
  publisher={Springer}
}

@inproceedings{wang2018reluval,
  title={Formal Security Analysis of Neural Networks using Symbolic Intervals},
  author={Wang, Shiqi and Pei, Kexin and Whitehouse, Justin and Yang, Junfeng and Jana, Suman},
  booktitle={27th USENIX Security Symposium (USENIX Security 18)},
  pages={1599--1614},
  year={2018}
}

@article{bunel2020branch,
  title={Branch and Bound for Piecewise Linear Neural Network Verification},
  author={Bunel, Rudy and Lu, Jingyue and Turkaslan, Ilker and Torr, Philip H. S. and Kohli, Pushmeet and Kumar, M. Pawan},
  journal={Journal of Machine Learning Research},
  volume={21},
  number={42},
  pages={1--39},
  year={2020}
}

@inproceedings{everett2021efficient,
  title={Efficient Reachability Analysis of Closed-Loop Systems with Neural Network Controllers},
  author={Everett, Michael and Habibi, Golnaz and How, Jonathan P.},
  booktitle={2021 IEEE International Conference on Robotics and Automation (ICRA)},
  year={2021},
  doi={10.1109/ICRA48506.2021.9561348}
}

@article{everett2021reachability,
  title={Reachability Analysis of Neural Feedback Loops},
  author={Everett, Michael and Habibi, Golnaz and Sun, Chuangchuang and How, Jonathan P.},
  journal={IEEE Access},
  volume={9},
  year={2021},
  doi={10.1109/ACCESS.2021.3133370}
}

@article{xiang2021reachable,
  title={Reachable Set Estimation for Neural Network Control Systems: A Simulation-Guided Approach},
  author={Xiang, Weiming and Tran, Hoang-Dung and Yang, Xiaodong and Johnson, Taylor T.},
  journal={IEEE Transactions on Neural Networks and Learning Systems},
  volume={32},
  number={5},
  pages={1821--1830},
  year={2021}
}

@inproceedings{ghosh2018verifying,
  title={Verifying Controllers Against Adversarial Examples with {Bayesian} Optimization},
  author={Ghosh, Shromona and Berkenkamp, Felix and Ranade, Gireeja and Qadeer, Shaz and Kapoor, Ashish},
  booktitle={2018 IEEE International Conference on Robotics and Automation (ICRA)},
  pages={7306--7313},
  year={2018},
  doi={10.1109/ICRA.2018.8460635}
}

@article{deshmukh2017testing,
  title={Testing Cyber-Physical Systems through {Bayesian} Optimization},
  author={Deshmukh, Jyotirmoy and Horvat, Marko and Jin, Xiaoqing and Majumdar, Rupak and Prabhu, Vinayak S.},
  journal={ACM Transactions on Embedded Computing Systems},
  volume={16},
  number={5s},
  pages={170:1--170:18},
  year={2017},
  doi={10.1145/3126521}
}

@article{chen2024end,
  title={End-to-End Autonomous Driving: Challenges and Frontiers},
  author={Chen, Li and Wu, Penghao and Chitta, Kashyap and Jaeger, Bernhard and Geiger, Andreas and Li, Hongyang},
  journal={IEEE Transactions on Pattern Analysis and Machine Intelligence},
  volume={46},
  number={12},
  pages={10164--10183},
  year={2024},
  doi={10.1109/TPAMI.2024.3435937}
}

@article{tampuu2020survey,
  title={A Survey of End-to-End Driving: Architectures and Training Methods},
  author={Tampuu, Ardi and Matiisen, Tambet and Semikin, Maksym and Fishman, Dmytro and Muhammad, Naveed},
  journal={IEEE Transactions on Neural Networks and Learning Systems},
  year={2020},
  doi={10.1109/TNNLS.2020.3043505}
}

@article{kroemer2021review,
  title={A Review of Robot Learning for Manipulation: Challenges, Representations, and Algorithms},
  author={Kroemer, Oliver and Niekum, Scott and Konidaris, George},
  journal={Journal of Machine Learning Research},
  volume={22},
  number={30},
  pages={1--82},
  year={2021}
}

@inproceedings{zhang2018efficient,
  title={Efficient Neural Network Robustness Certification with General Activation Functions},
  author={Zhang, Huan and Weng, Tsui-Wei and Chen, Pin-Yu and Hsieh, Cho-Jui and Daniel, Luca},
  booktitle={Advances in Neural Information Processing Systems},
  volume={31},
  pages={4939--4948},
  year={2018}
}

@article{gowal2018effectiveness,
  title={On the Effectiveness of Interval Bound Propagation for Training Verifiably Robust Models},
  author={Gowal, Sven and Dvijotham, Krishnamurthy and Stanforth, Robert and Bunel, Rudy and Qin, Chongli and Uesato, Jonathan and Arandjelovic, Relja and Mann, Timothy and Kohli, Pushmeet},
  journal={arXiv preprint arXiv:1810.12715},
  year={2018}
}

@article{brockman2016openai,
  title={OpenAI Gym},
  author={Brockman, Greg and Cheung, Vicki and Pettersson, Ludwig and Schneider, Jonas and Schulman, John and Tang, Jie and Zaremba, Wojciech},
  journal={arXiv preprint arXiv:1606.01540},
  year={2016}
}

@book{rasmussen2006gaussian,
  title={Gaussian Processes for Machine Learning},
  author={Rasmussen, Carl Edward and Williams, Christopher K. I.},
  publisher={MIT Press},
  year={2006}
}

@inproceedings{bryan2005active,
  title={Active Learning for Identifying Function Threshold Boundaries},
  author={Bryan, Brent and Schneider, Jeff and Nichol, Robert C. and Miller, Christopher J. and Genovese, Christopher R. and Wasserman, Larry},
  booktitle={Advances in Neural Information Processing Systems},
  volume={18},
  pages={163--170},
  year={2005}
}

@inproceedings{gotovos2013active,
  title={Active Learning for Level Set Estimation},
  author={Gotovos, Alkis and Casati, Nathalie and Hitz, Gregory and Krause, Andreas},
  booktitle={Proceedings of the International Joint Conference on Artificial Intelligence (IJCAI)},
  pages={1344--1349},
  year={2013}
}

@article{levine2016end,
  title={End-to-End Training of Deep Visuomotor Policies},
  author={Levine, Sergey and Finn, Chelsea and Darrell, Trevor and Abbeel, Pieter},
  journal={Journal of Machine Learning Research},
  volume={17},
  number={39},
  pages={1--40},
  year={2016}
}

@article{jaulin1993set,
  title={Set inversion via interval analysis for nonlinear bounded-error estimation},
  author={Jaulin, Luc and Walter, Eric},
  journal={Automatica},
  volume={29},
  number={4},
  pages={1053--1064},
  year={1993},
  doi={10.1016/0005-1098(93)90106-4}
}

@inproceedings{donze2009parameter,
  title={Parameter Synthesis for Hybrid Systems with an Application to {Simulink} Models},
  author={Donz{\'e}, Alexandre and Krogh, Bruce and Rajhans, Akshay},
  booktitle={Hybrid Systems: Computation and Control (HSCC)},
  series={LNCS},
  volume={5469},
  pages={165--179},
  year={2009},
  publisher={Springer},
  doi={10.1007/978-3-642-00602-9_12}
}


\clearpage
\appendix
\section{Appendix\\[0.2em]}
This appendix provides additional details on the neural network architectures, the dynamics, and the experimental results.

\noindent
\subsection{Neural Network Architectures}
For all four benchmarks, the world model and controller architectures are
identical except for the final activation of the controller.
The controller is trained with reinforcement learning.
All the controller's parameters are provided below in Table~\ref{tab:ctrl_arch}, and the world model decoder parameters are provided in Table~\ref{tab:wm_arch}.

\begin{table}[H]
\centering
\caption{World Model Decoder Architecture (all benchmarks).}
\label{tab:wm_arch}
\begin{tabular}{l}
\hline
\textbf{Decoder $\hat P$: state $\in \mathbb{R}^2$ $\to$ image $\in \mathbb{R}^{1\times96\times96}$} \\
\hline
Dense $\to$ 32, ReLU \\
Dense $\to$ 64, ReLU \\
Dense $\to$ $3 \times 12 \times 12$, ReLU \\
Reshape $\to$ $3 \times 12 \times 12$ \\
ConvTranspose: $3 \to 4$, kernel $4\times4$, stride 2, padding 1, ReLU \\
ConvTranspose: $4 \to 8$, kernel $4\times4$, stride 2, padding 1, ReLU \\
ConvTranspose: $8 \to 1$, kernel $4\times4$, stride 2, padding 1, SatLin \\
\hline
\end{tabular}
\end{table}

\begin{table}[H]
\centering
\caption{Image-Based Controller Architecture.}
\label{tab:ctrl_arch}
\begin{tabular}{l}
\hline
\textbf{All controllers: image $\in \mathbb{R}^{1\times96\times96}$ $\to$ action $\in \mathbb{R}$} \\
\hline
Conv: $1 \to 4$, kernel $4\times4$, stride 2, padding 1, ReLU \\
Conv: $4 \to 1$, kernel $4\times4$, stride 2, padding 1, ReLU \\
Flatten $\to$ 576 \\
Dense $\to$ 64, ReLU \\
Dense $\to$ 1, \textit{Sigmoid} (CartPole/Brake) \\
/ \textit{Tanh} (MountainCar, Pendulum) \\
\hline
\end{tabular}
\end{table}

\noindent
\subsection{Dynamics for Four Case Studies}

We next give the dynamics of all four benchmarks.
\paragraph{CartPole (Gym)}
The state is $s = (x, \dot{x}, \theta, \dot{\theta}) \in \mathbb{R}^4$,
where $x$ is the cart position and $\theta$ is the pole angle.
The control input $u$ represents the applied force.

The continuous-time dynamics are given by:
\[
\ddot{x} = \frac{F + m_p \sin\theta \bigl(l \dot{\theta}^2 + g \cos\theta\bigr)}
{m_c + m_p \sin^2\theta},
\]
\[
\ddot{\theta} = \frac{-F \cos\theta - m_p l \dot{\theta}^2 \cos\theta \sin\theta
- (m_c + m_p) g \sin\theta}
{l (m_c + m_p \sin^2\theta)},
\]
where $F = F_{\max} u$.
We discretise the system with time step $\Delta t$:
\[
x_{t+1} = x_t + \Delta t \,\dot{x}_t, \quad
\dot{x}_{t+1} = \dot{x}_t + \Delta t \,\ddot{x}_t,
\]
\[
\theta_{t+1} = \theta_t + \Delta t \,\dot{\theta}_t, \quad
\dot{\theta}_{t+1} = \dot{\theta}_t + \Delta t \,\ddot{\theta}_t.
\]

\paragraph{MountainCar (Gym)}
The state is $s = (p, v) \in \mathbb{R}^2$, where $p$ is position and $v$ is velocity.
The control input $u$ is the throttle force.

The discrete-time dynamics are:
\[
v_{t+1} = v_t + 0.001\,u_t - 0.0025 \cos(3 p_t),
\]
\[
p_{t+1} = p_t + v_{t+1}.
\]

The state is clipped to the admissible ranges
$p \in [-1.2, 0.6]$ and $v \in [-0.07, 0.07]$.

\paragraph{Pendulum (Gym)}
The state is $s = (\theta, \dot{\theta}) \in \mathbb{R}^2$,
where $\theta$ is the angle and $\dot{\theta}$ is the angular velocity.
The control input $u$ is the applied torque.

The dynamics follow:
\[
\ddot{\theta} = -\frac{3g}{2l} \sin(\theta + \pi) + \frac{3}{ml^2} u,
\]
and are discretised as:
\[
\dot{\theta}_{t+1} = \dot{\theta}_t + \Delta t \,\ddot{\theta}_t,
\]
\[
\theta_{t+1} = \theta_t + \Delta t \,\dot{\theta}_{t+1}.
\]




\paragraph{Advanced Emergency Braking System (AEBS)}
The state is $s = (d, v) \in \mathbb{R}^2$, where $d$ is the relative distance to the
leading vehicle and $v$ is the ego velocity; the control input $u \in [0,1]$ represents
the normalized braking intensity, with $u = 0$ meaning no braking and $u = 1$ full
braking. At each step the vision-based controller observes an image of the road scene,
infers the braking command $u_t = C(I_t)$, and the vehicle decelerates accordingly.

The continuous-time dynamics describe a vehicle closing on a stationary (or slower)
leading vehicle while braking: the gap $d$ shrinks at the ego velocity $v$, and $v$
decreases in proportion to the applied braking effort. Discretizing with a fixed step
$\Delta t$ yields
\[
d_{t+1} = d_t - \Delta t \, v_t,
\]
\[
v_{t+1} = v_t - \Delta t \, a_{\max} \, u_t,
\]
where $a_{\max}$ is the maximum deceleration achievable at full braking, so that the
effective deceleration $a_{\max} u_t$ scales linearly with the command. The velocity is
clipped to $v \ge 0$, as the model does not capture reversing.

The safety objective is to avoid a collision with the leading vehicle throughout the
horizon. A collision occurs whenever the relative distance is non-positive, giving the
unsafe set
\[
X_{\mathrm{unsafe}} = \{(d,v) \mid d \le 0\}.
\]
Verification therefore certifies that, starting from an initial set of gap–velocity
configurations, the closed-loop trajectory brings the vehicle to a stop with $d > 0$
maintained at every step.

\paragraph{Braking System (4-D)}
The four-dimensional variant used in Section~\ref{subsec:partition} adds the brake gain
$k_{\rm brake}$ and the base deceleration $a_0$ to the state,
$s = (d, v, k_{\rm brake}, a_0) \in \mathbb{R}^4$. Both are constant along a trajectory
and model the uncertainty in road friction and brake wear. The controller output
$C(I_t) \in [-1,1]$ is mapped to the braking intensity
$u_t = \mathrm{clip}\big(\tfrac{1}{2}(C(I_t)+1), 0, 1\big)$, and the dynamics are
\[
d_{t+1} = d_t - \Delta t \, v_t, \qquad
v_{t+1} = \max\!\big(0,\; v_t - \Delta t \,(k_{\rm brake}\, u_t + a_0)\big),
\]
\[
k_{{\rm brake},t+1} = k_{{\rm brake},t}, \qquad a_{0,t+1} = a_{0,t},
\]
with $\Delta t = 0.1$\,s and $T = 20$ steps, so that the horizon covers a full stop.
The initial set is
\[
S_0 = [6,8]\,\text{m} \times [6,6.4]\,\text{m/s} \times [8,10]\,\text{m/s}^2 \times [0.3,0.55]\,\text{m/s}^2 ,
\]
and the two-dimensional Braking System is its slice at the nominal values
$k_{\rm brake} = 9.0$\,m/s$^2$ (about $0.9\,g$ at full braking) and $a_0 = 0.42$\,m/s$^2$.
The world model and the controller are the same networks as in the two-dimensional
benchmark: the decoder takes only $(d, v)$, so the camera does not see $k_{\rm brake}$ or
$a_0$, which reach the image only through the distance in later frames.
The specification requires the vehicle never to collide and to stop no farther than
$d_{\max} = 3.5$\,m from the leading vehicle,
\[
\rho_\varphi(\tau_{\rm wm}(s_0,T)) = \min\Big(\min_{0\le t\le T} d_t,\; d_{\max} - d_T\Big),
\]
so that both under-braking (collision) and over-braking (stopping too early) violate it.
Since $u_t$, $k_{\rm brake}$, and $a_0$ are non-negative, the deceleration is monotone in each
of them, and the interval step takes the lower velocity bound from the upper bounds of
$u_t$, $k_{\rm brake}$, and $a_0$, and the upper velocity bound from their lower bounds.
On $S_0$, about $80\%$ of $50{,}176$ uniformly sampled initial states satisfy $\varphi$.

\noindent
\subsection{Pixel-level Interval Examples}

\begin{figure}[H]
 \begin{center}
 \includegraphics[width=1.25\linewidth, trim={120 0 0 0}, clip]
 {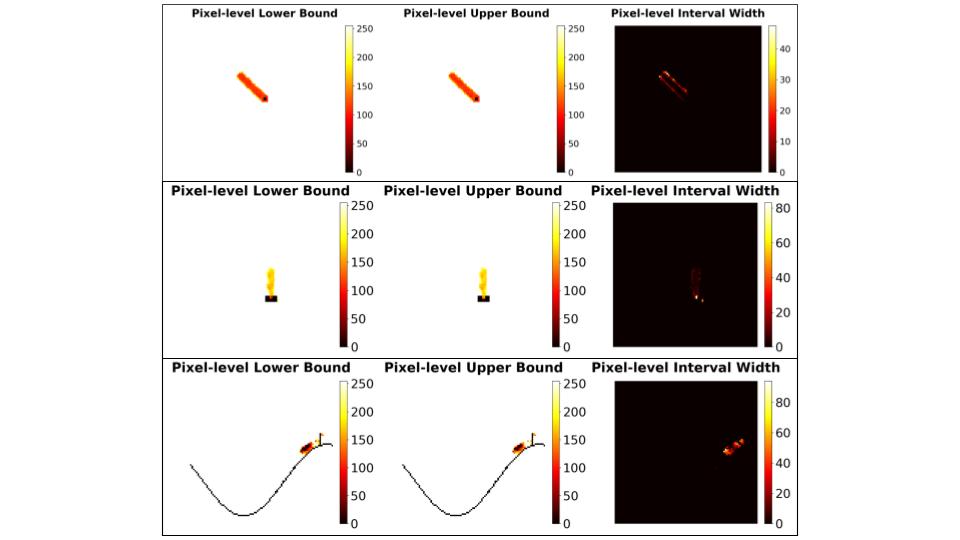}
 
 \caption{Pixel-level DWM bounds (lower bound, upper bound, width) for Pendulum (top), CartPole (middle), and MountainCar (bottom). The uncertainty concentrates on task-relevant regions.}
 \label{fig: pixel_level_bound}
 \end{center}

\end{figure}

Figure~\ref{fig: pixel_level_bound} visualizes the pixel-level reachable
sets produced by the proposed DWM-based verification pipeline under
identical initial state intervals.
Each row corresponds to a different case study, namely the Advanced
Emergency Braking System (top), Pendulum (middle), and MountainCar (bottom).
For each case, we show the lower bound image, upper bound image, and the
corresponding pixel-wise interval width.

Given an initial Star set $\mathcal{S}_0$, the decoder reachability
$R_{\hat P}(\mathcal{S}_0)$ produces an ImageStar representation,
which encodes a set of images with shared predicate variables.
The lower and upper bound images are obtained by solving two linear
programs per pixel:
\[
\underline{I}(p) = \min_{\alpha \in \mathcal{A}} I(p,\alpha), \quad
\overline{I}(p) = \max_{\alpha \in \mathcal{A}} I(p,\alpha),
\]
where $p$ indexes pixel locations and $\mathcal{A}$ denotes the feasible
predicate set of the ImageStar.

The interval width image is defined as:
\[
W(p) = \overline{I}(p) - \underline{I}(p),
\]
which characterizes the uncertainty induced by the initial state set
through the decoder.

\end{document}